\documentclass[11pt]{article}

\usepackage[final]{acl}

\usepackage{times}
\usepackage{latexsym}
\usepackage[T1]{fontenc}
\usepackage[utf8]{inputenc}
\usepackage{microtype}
\usepackage{inconsolata}
\usepackage{graphicx}
\usepackage{tikz}
\usetikzlibrary{positioning}

\usepackage{booktabs}
\usepackage{multirow}
\usepackage{amsmath}
\usepackage{amsfonts}
\usepackage{mathtools}
\usepackage{algorithm}
\usepackage{algorithmic}
\usepackage{enumitem}

\title{Rethinking Calibration for High-Sparsity LLM Pruning: From Single Sources to Capability Coverage}

\author{
 \textbf{Hu Xu\textsuperscript{1,2}\thanks{\ \ Work done during internship at JD.}},
 \textbf{Zhaolong Xing\textsuperscript{2}},
 \textbf{Congcong Liu\textsuperscript{2}},
 \textbf{Jiaxing Wang\textsuperscript{2}},
\\
 \textbf{Zhida Jiang\textsuperscript{2}},
 \textbf{Junshi Huang\textsuperscript{2}},
 \textbf{Zhen Chen\textsuperscript{2}\thanks{\ \ Jianfeng Xu and Zhen Chen are co-corresponding authors.}}
 \textbf{Jianfeng Xu\textsuperscript{1}\footnotemark[2]}
\\
\\
 \textsuperscript{1}Shanghai Jiao Tong University,
 \textsuperscript{2}JD.com
\\
 \small{
   \texttt{\{xuhu6736, xujf\}@sjtu.edu.cn},
   \texttt{\{chenzhen48, xingzhaolong1\}@jd.com}
 }
}

\begin{document}
\maketitle

\begin{abstract}
% Post-training LLM pruning relies on a small calibration set, usually chosen heuristically and judged by WikiText perplexity or by mean accuracy on general-domain commonsense benchmarks. This single-axis view hides capability-specific effects: at 60\% sparsity, strategies separated by $2.85$ points in averaged commonsense accuracy diverge by $51.9$ points in code retention. We treat calibration as a \emph{capability-coverage} problem. \textbf{Balanced}, a uniform four-way mixture, tests the principle on real corpora, and its margin over C4 grows from $+4.6$ points at 30\% sparsity to $+18.8$ at 60\%. When aligned corpora are unavailable, \textbf{IGSP} approximates that mixture from the base model and the evaluation taxonomy alone. On Qwen2.5-7B, IGSP exceeds Balanced on both the main suite and held-out benchmarks. Capability coverage is a first-order design variable for high-sparsity pruning.

Post-training LLM pruning relies on a small calibration set, usually chosen heuristically and judged by WikiText perplexity or by mean accuracy on general-domain commonsense benchmarks. This single-axis view hides capability-specific effects: at 60\% sparsity, strategies separated by $2.85$ points in averaged commonsense accuracy diverge by $51.9$ points in code retention, and across 15 sources no single one preserves all capabilities. We treat calibration as a \emph{capability-coverage} problem. \textbf{Balanced}, a uniform four-way mixture under a fixed budget, tests the principle on real corpora: it outperforms every evaluated single source, and its margin over C4 grows from $+4.6$ points at 30\% sparsity to $+18.8$ at 60\%. The advantage holds across model scale and a second model family, and transfers to held-out benchmarks disjoint from calibration construction. When aligned corpora are unavailable, \textbf{IGSP} approximates that mixture from the base model and the evaluation taxonomy alone, outperforming prior self-generation methods by $7.23$--$12.79$ points on the LLaMA main suite. On Qwen2.5-7B, IGSP exceeds Balanced on both the main suite and held-out benchmarks. Capability coverage is a first-order design variable for high-sparsity pruning.

\end{abstract}

\section{Introduction}
% v3 changes vs v2:
%  * Spearman paragraph re-tiered per E2 (BH-FDR + LOSO): only the
%    Aggregation--General association survives correction; the opposing
%    Distortion associations are reported as exploratory. The v2 wording
%    "(all p<0.05)" is removed.
%  * OIT sentence softened from "the only two ... that consistently
%    correlate" to directionally-stable wording.
%  * Contributions rewritten per plan §5 (principle-first framing);
%    contribution 3 states the coverage/selection split.

\begin{figure*}[htbp]
  \centering
  \includegraphics[width=0.96\textwidth]{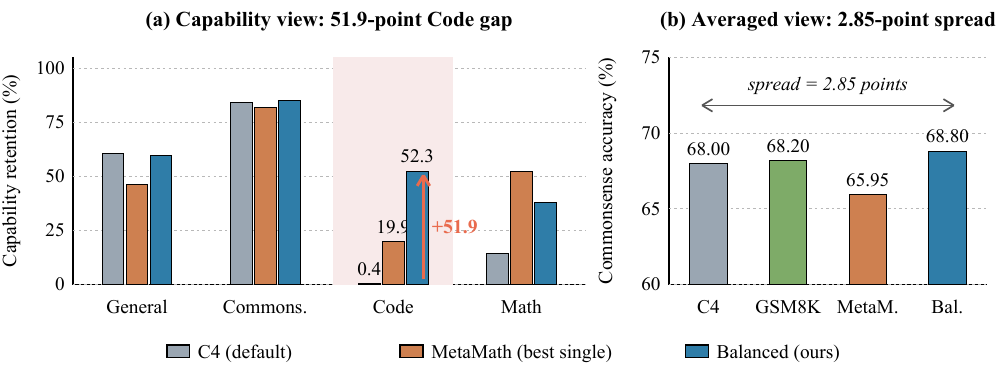}
  \caption{Capability-decomposed evaluation reveals a $51.9$-point Code gap that a $2.85$-point commonsense average conceals. Results are for LLaMA-3.1-8B pruned by SparseGPT to 60\% sparsity. (a) Code retention rises from $0.4$ with C4 and $19.9$ with MetaMath, the strongest single source, to $52.3$ with Balanced. (b) The same strategies span $2.85$ points in averaged commonsense accuracy (PIQA and ARC-Easy).}
  \label{fig:hook}
\end{figure*}

Post-training pruning can remove 50--60\% of an LLM's weights without retraining, yet its outcome may depend on as few as 128--256 calibration sequences. SparseGPT~\cite{frantar2023sparsegpt} uses these samples to approximate layer-wise Hessians, whereas Wanda~\cite{wanda} uses them to estimate activation magnitudes. The calibration set therefore influences which weights survive, but its composition is often treated as a minor implementation detail, with C4 and Wikipedia serving as common defaults.

Recent work has begun to question these defaults. \citet{williams2024impact} compare six calibration sources for quantization and pruning and report only modest changes in accuracy averaged across general-domain benchmarks, while \citet{bandari2024c4} investigate whether C4 is an optimal pruning default. A parallel line removes the need for external data: Self-Cal~\cite{williams2025self} generates calibration text from the base model, and SGS~\cite{ji2025beware} filters such text using vocabulary entropy and a target perplexity distribution. Despite their different constructions, these approaches share two evaluation choices: they assess calibration through \emph{averaged} performance on general-domain benchmarks and draw calibration samples from a \emph{single} corpus or self-generation distribution. This protocol cannot reveal whether a source preserves one capability by sacrificing another.

We find that the apparent robustness to calibration choice is an averaging artifact (Figure~\ref{fig:hook}). Among single sources, C4 retains $60.7\%$ of General capability but only $0.4\%$ of Code, whereas MetaMath retains $52.2\%$ of Math but only $46.2\%$ of General. Across $n{=}15$ sources, calibration perplexity is positively associated with General retention and negatively associated with Math and Code retention (\S\ref{sec:correlation}). Averaging hides this trade-off through incomplete capability coverage and through cancellation of opposite-signed changes. None of the evaluated single sources resolves it.

\textbf{Balanced} tests capability coverage when aligned corpora exist; Information-Guided Self-Calibration for Pruning (\textbf{IGSP}) approximates the same principle when they do not. Balanced draws uniformly from General, Commonsense, Code, and Math pools under the same 128-sequence budget as every baseline. IGSP uses only the evaluation taxonomy and the base model: it builds dimension-specific pools and selects low-redundancy samples inside per-dimension perplexity bands (Figure~\ref{fig:igsp_pipeline}).

Three conclusions follow. Coverage gains grow with sparsity on LLaMA-3.1-8B and Qwen2.5-7B. On Qwen2.5-7B, IGSP exceeds Balanced on the main suite and on held-out benchmarks, and the same SparseGPT ordering holds at 70B. The calibration spread is wider under SparseGPT than under Wanda.

Our main contributions are threefold:
\begin{enumerate}
    \item \textbf{A capability-decomposed diagnosis of calibration evaluation.} We show that prevailing single-axis evaluation hides calibration effects through two distinct mechanisms---incomplete capability coverage and genuine cross-dimension cancellation---and identify opposite-sign associations between calibration perplexity and General versus Math/Code retention across $n{=}15$ sources (\S\ref{sec:single_axis}--\S\ref{sec:correlation}).
    \item \textbf{Capability coverage as a design principle, validated across regimes.} Under a fixed budget, a capability-balanced mixture outperforms every evaluated single source. On LLaMA-3.1-8B and Qwen2.5-7B its margin over C4 grows with sparsity; the advantage persists at 70B, and on LLaMA-3.1-8B it is more than four times larger under SparseGPT than under Wanda (\S\ref{sec:multi_source}, \S\ref{sec:wanda_saturation}).
    \item \textbf{Approximating the principle without aligned corpora.} IGSP instantiates capability coverage through stratified self-generation with perplexity-range and redundancy control. On LLaMA-3.1-8B at SparseGPT 60\%, coverage alone exceeds Self-Cal and SGS, and information-guided selection adds a further $7.9$ points. On Qwen2.5-7B, IGSP's SparseGPT Total exceeds Balanced on both the main suite and held-out benchmarks (Table~\ref{tab:cross_family}, \S\ref{sec:igsp_automation}).
\end{enumerate}

\section{Related Work}
% v3 changes vs v2 (plan §4):
%  * NEW paragraph situating he2025preserving (COLA); shin2024rethinking
%    placed at the front of the self-generation lineage.
%  * "robust associations" claim re-tiered per E2 (only Aggregation--General
%    survives BH-FDR).
%  * 2026-10-10 rewrite: five paragraphs -> three, ~610 -> ~400 words,
%    following the style of Williams & Aletras (ACL'24) and Self-Cal
%    (NAACL'25): survey first, distinction in one or two closing sentences.
%    The four-point COLA comparison is compressed to two points (per-dimension
%    high-sparsity evaluation; regime dependence), the rest lives in §5.
%  * 2026-10-10: absorbed Methodology's "Relation to prior work" paragraph
%    into the self-generated calibration paragraph (entropy-vs-Aggregation
%    contrast with numbers); Methodology no longer carries related-work text.

\paragraph{Calibration data for post-training compression.}
Post-training pruning~\cite{frantar2023sparsegpt,wanda} and quantization~\cite{frantar2022gptq,lin2024awq} methods rely on a small unlabeled calibration set, conventionally sampled at random from web text. A recent line of work examines how this set should be constructed: \citet{williams2024impact} compare six sources at sparsity $\leq 50\%$ and report modest differences in averaged accuracy, \citet{bandari2024c4} question whether C4 is an optimal pruning default, and \citet{kurz2024investigating} extend the question to language-specific calibration for multilingual LLMs. Closest to our work, \citet{he2025preserving} analyze how compositional properties and domain correspondence of calibration data affect the capabilities of compressed LLMs, and curate calibration sets through domain matching and activation-space diversity (COLA). These studies evaluate sources through averaged or domain-level outcomes at moderate sparsity. Compressed-model evaluation has expanded beyond a single aggregate, including factual knowledge~\cite{hoang2023compressed}, summarization~\cite{chrysostomou2024investigating}, and predictive confidence~\cite{metzler2024quantization}. A narrow evaluation axis can conceal capability trade-offs of the same kind. We measure capability-specific retention at high sparsity, where averaging hides large trade-offs (\S\ref{sec:single_axis}), and show that calibration conclusions are regime-dependent: C4, competitive for code retention at moderate sparsity in the COLA analysis, collapses in code retention at 60\% sparsity here. Our design stays simple: uniform capability-balanced mixing under a fixed budget, without activation-space computation.

\paragraph{Self-generated calibration.}
\citet{shin2024rethinking} first showed that text self-generated by the dense model mitigates the overfitting of reconstruction-based pruning to small calibration sets. Self-Cal~\citep{williams2025self} systematizes this idea, matching or exceeding Wikipedia and C4 on standard benchmarks, and SGS~\citep{ji2025beware} further filters self-generated text toward a target perplexity distribution. All three construct a \emph{single global} generation distribution, whereas IGSP generates capability-stratified pools and balances perplexity across dimensions, motivated by the capability-specific perplexity regimes revealed by our analysis (\S\ref{sec:correlation}). The filtering signal also differs: SGS applies vocabulary-level entropy, but our analysis identifies phrase-level aggregation as a substantially stronger correlate of General retention ($|\rho|{=}0.79$ vs.\ ${<}0.2$), which IGSP therefore controls through low-redundancy selection.

\section{Methodology}
\label{sec:method}
% v3 changes vs v2:
%  * §3.2 hierarchy paragraph re-tiered per E2 (BH-FDR + LOSO).
%  * §3.3 Stage 2 rewritten per plan §5: PPL bands are derived from the
%    PPL distributions of the PUBLISHED SOURCE CORPORA (not the evaluation
%    tasks), fixing the circularity description in v2 ("empirical PPL
%    distribution of the corresponding evaluation tasks") that invited the
%    "benchmark-informed" critique. Verified against the code: bands are
%    computed once from source-corpus PPL under the unpruned model and
%    frozen across pruners/sparsity/seeds (v3_coling/derive_ppl_bands.py).
%  * 2026-10-10: held-out transfer sentence enabled (E3 re-run landed).
%  * 2026-10-10: removed the "Relation to prior work" paragraph; the
%    IGSP-vs-SGS/Self-Cal distinctions now live in Related Work
%    (self-generated calibration paragraph), per standard practice.

\begin{figure}[t]
  \centering
  \includegraphics[width=\columnwidth]{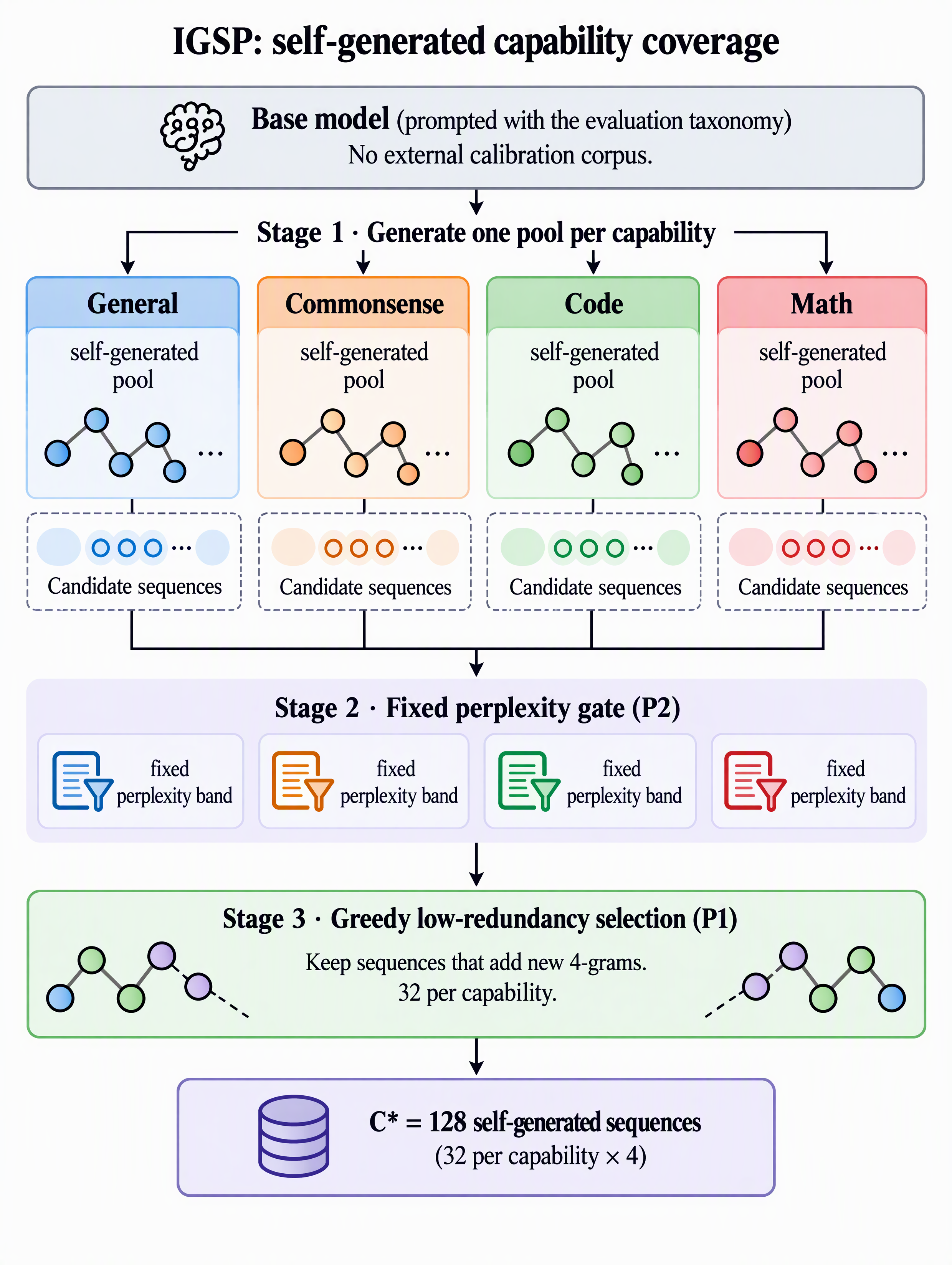}
  \caption{IGSP constructs a self-generated calibration set without an external corpus. Stage~1 prompts the base model to write one pool per capability. Stage~2 keeps samples inside a fixed perplexity band and then selects low-redundancy sequences until $C^{*}$ has 128 sequences.}
  \label{fig:igsp_pipeline}
\end{figure}

Balanced tests capability coverage when aligned corpora exist. IGSP approximates the same principle when they do not, using only the base model and the evaluation taxonomy. This section defines the retention metric, the information profile that motivates the selection rule, and the IGSP procedure.

\subsection{Multi-dimensional Capability Retention}
\label{sec:multidim-metric}

Let $m_0$ denote the unpruned model and $m$ a pruned variant. Evaluation tasks are partitioned into four dimensions $\mathcal{D}=\{\textsc{General},\textsc{Commonsense},\textsc{Code},\textsc{Math}\}$, with $\mathcal{T}_d$ the benchmarks assigned to dimension $d$ (Appendix~\ref{app:tasks_taxonomy}). For each task $t$ with metric $\mathrm{score}(m,t)\in[0,1]$, we define the \emph{relative retention}
\begin{equation}
\tilde{y}_{m,t} \;=\; \frac{\mathrm{score}(m,t)}{\mathrm{score}(m_0,t) + \varepsilon},
\end{equation}
and aggregate to dimension- and overall-level scores:
\[
S_{m,d} = \tfrac{1}{|\mathcal{T}_d|}\!\sum_{t\in\mathcal{T}_d}\!\tilde{y}_{m,t},
\qquad
S^{\text{total}}_m = \tfrac{1}{|\mathcal{D}|}\!\sum_{d\in\mathcal{D}}\!S_{m,d}.
\]
This normalisation makes comparisons consistent across tasks, sparsity levels, and pruning methods. Multi-seed aggregation and edge cases are detailed in Appendix~\ref{app:multidim-eval}.

\subsection{Information Profile of Calibration Data}
\label{sec:info-profile}

We characterise each calibration set $C$ via an information profile~\cite{xu2023foundations,xu2024research,xu2025general}:
\begin{equation}
m(C) = \big[\,\mathrm{Vol},\,\mathrm{S},\,\mathrm{Var},\,\mathrm{G},\,\mathrm{A},\,\mathrm{D}_M,\,\mathrm{M}_{I_0,M}\,\big](C),
\end{equation}
where $\mathrm{Vol}$ is the token budget, $\mathrm{S}$ (Scope) and $\mathrm{Var}$ (Variety) capture coverage breadth, $\mathrm{G}$ (Granularity) measures partition fineness, $\mathrm{A}$ (Aggregation) is the 4-gram redundancy, $\mathrm{D}_M$ (Distortion) is the perplexity under the base model, and $\mathrm{M}_{I_0,M}$ (Mismatch) is the Min-K\%++ score~\cite{shi2024min}. Operational definitions are in Appendix~\ref{app:oit-metrics}.

A correlation study on $n{=}15$ calibration sets identifies Aggregation and Distortion as the two signals associated with capability retention; the remaining descriptors are diagnostic only. We therefore optimise only those two in IGSP. Coefficients, FDR correction, and leave-one-source-out checks are in \S\ref{sec:correlation}.

\subsection{IGSP: Information-Guided Self-Calibration for Pruning}
\label{sec:igsp-main}

IGSP constructs a calibration set $C^*$ of fixed budget $B$ that spans the perplexity regimes required for cross-capability coverage. It uses only the base model and the evaluation protocol, without requiring access to the model's original pre-training corpus.

\paragraph{Design principles.}
\textbf{P1 (Aggregation).} Calibration data with high 4-gram redundancy biases activation estimation toward repeated patterns; we greedily select samples that maximise unique $n$-gram coverage.
\textbf{P2 (Distortion balance).} Since no single perplexity level supports all capabilities, the calibration set must include data from both high- and low-perplexity regimes, allocated across capability dimensions.

\paragraph{Running example.}
Figure~\ref{fig:igsp_pipeline} shows the procedure. For a budget of 128 sequences, uniform allocation asks for 32 Math sequences. Stage 1 fills a Math pool by prompting the base model to generate grade-school word problems. Stage 2 keeps a candidate only if its perplexity under the unpruned model lies in the Math band $[2.0,\,4.0]$ (Table~\ref{tab:ppl_bands}), then greedily prefers candidates that add new 4-grams until 32 sequences are chosen. General, Commonsense, and Code follow the same steps with their own bands. The four slices are concatenated into $C^*$. Each band is a fixed gate, computed once from published source corpora and not tuned on post-pruning retention.

\paragraph{Construction.}

\textbf{Stage 1 (capability-stratified pools).} For each $d\in\mathcal{D}$, we build a candidate pool $P_d$ aligned with that capability, either by partitioning existing corpora (e.g., GSM8K and MetaMath for Math) or by prompting the base LLM with dimension-specific templates (e.g., ``Generate a grade-school math word problem with solution''). The full pool is $P=\bigcup_d P_d$. This stage implements capability coverage.

\textbf{Stage 2 (per-dimension selection).} From each $P_d$, we select $b_d=\alpha_d B$ samples (uniformly by default, $\alpha_d{=}1/|\mathcal{D}|$) while greedily minimising Aggregation subject to a dimension-specific perplexity band $[\tau_d^{\text{lo}},\tau_d^{\text{hi}}]$. Each band is derived \emph{once}, before any pruning experiment, from the perplexity distribution of the \emph{published source corpora} assigned to dimension $d$ under the unpruned model (e.g., the Math band spans the observed perplexities of GSM8K-train and MetaMath; see Appendix~\ref{app:igsp_ppl_bands} for values and provenance). The bands are fixed gates: they stay constant across candidate pools, pruners, sparsity levels, and seeds, and they are not tuned on post-pruning retention. This stage implements P1 and P2. Calibration construction is thus separated from evaluation: no test input from the evaluation suite is used to set any selection hyperparameter.
We assess transfer to held-out capability benchmarks in \S\ref{sec:heldout}.
The procedure runs in $O(B\cdot|P|)$ time; pseudocode is provided in Appendix~\ref{app:igsp}.

\section{Experimental Setup}
\label{sec:exp}
% v3 changes vs v2:
%  * I4: "eight benchmarks" -> accurate count (nine configurations;
%    LAMBADA-Std and LAMBADA-OpenAI enter the General average as two
%    separate tasks).
%  * I5: display names unified (Balanced / Math-heavy / Code-heavy) with a
%    footnote mapping to the configuration identifiers used in code.
%  * I12: Table 1 caption now states that MetaMath is the strongest of all
%    15 evaluated single sources (full 15-source table in the appendix).
%  * mix_balanced recipe: CONFIRMED by the author (2026-10-08) as uniform
%    1:1:1:1 across the four capability pools. The local code's 52:0:38:38
%    mix without a commonsense pool (PATCHES.md patch 4) is a stale/incorrect
%    local variant and must NOT be used for any regenerated numbers.
%  * 2026-10-12: Qwen2.5-7B added. OPT-6.7B appendix removed.
%    Qwen IGSP rows include the 30\% sparsity setting.

\paragraph{Models and pruners.}
The main comparison is \textbf{LLaMA-3.1-8B} and \textbf{Qwen2.5-7B}. \textbf{LLaMA-3.1-70B} is a 60\% sparsity scale check in Appendix~\ref{app:llama70}.
We apply three post-training pruners at $\{30,40,50,60\}\%$ unstructured sparsity on the two main models, without subsequent fine-tuning. \textbf{Wanda}~\cite{wanda} and \textbf{SparseGPT}~\cite{frantar2023sparsegpt} use a calibration set, and their pruning hyperparameters are fixed across calibration strategies. \textbf{Magnitude}~\cite{han2015learning} ranks weights by absolute value and uses no calibration set; it is the calibration-free reference in Table~\ref{tab:main_results}. At 60\% sparsity its Total retention is $9.50$ on LLaMA-3.1-8B and $14.46$ on Qwen2.5-7B. The 70B Magnitude row, Total $35.50$, is in Appendix~\ref{app:llama70}. Table~\ref{tab:igsp_ablation} tracks the LLaMA-3.1-8B comparison across sparsity, including the IGSP ablations.

\begin{table*}[t]
    \centering
    \small
    \begin{tabular}{@{}llccccc@{\qquad}ccccc@{}}
            \toprule
            & & \multicolumn{5}{c}{\textbf{LLaMA-3.1-8B}} & \multicolumn{5}{c}{\textbf{Qwen2.5-7B}} \\
            \cmidrule(lr){3-7} \cmidrule(lr){8-12}
            Method & Calibration & $S_{\text{Gen}}$ & $S_{\text{Com}}$ & $S_{\text{Code}}$ & $S_{\text{Math}}$ & $S^{\text{tot}}$
                                  & $S_{\text{Gen}}$ & $S_{\text{Com}}$ & $S_{\text{Code}}$ & $S_{\text{Math}}$ & $S^{\text{tot}}$ \\
            \midrule
            Magnitude & None
                & 14.50 & 20.40 & 1.20 & 1.90 & 9.50
                & 2.21 & 54.82 & 0.00 & 0.80 & 14.46 \\
            \midrule
            \multirow{10}{*}{Wanda}
            & C4
                & 41.72 & 80.63 & 1.31 & 12.37 & 34.01
                & 63.94 & 89.62 & 8.65 & 28.20 & 47.60 \\
            & Wikipedia
                & 38.75 & 80.13 & 6.60 & 13.94 & 34.86
                & 62.78 & 89.30 & 12.40 & 27.26 & 47.93 \\
            & GSM8K
                & 34.54 & 78.53 & 8.39 & 18.03 & 34.87
                & 56.58 & 90.05 & 31.55 & 49.92 & 57.02 \\
            & MetaMath
                & 35.26 & 79.95 & 6.67 & 17.85 & 34.93
                & 54.75 & 89.30 & 34.45 & 52.91 & 57.85 \\
            \cmidrule(lr){2-12}
            & Math-heavy
                & 40.90 & 77.98 & 7.49 & 14.52 & 35.22
                & 63.05 & 90.24 & 27.56 & 50.59 & 57.86 \\
            & Code-heavy
                & 39.03 & 80.50 & 14.72 & 14.01 & 37.06
                & 62.66 & 88.37 & 47.43 & 37.94 & 59.10 \\
            & Balanced
                & 40.89 & 79.77 & 14.65 & 15.75 & 37.76
                & 63.43 & 88.93 & 46.10 & 47.67 & 61.53 \\
            \cmidrule(lr){2-12}
            & Self-Cal
                & 38.93 & 81.37 & 12.79 & 13.32 & 36.60
                & 57.87 & 90.24 & 40.00 & 39.55 & 56.92 \\
            & SGS
                & 36.90 & 81.74 & 12.03 & 13.42 & 36.02
                & 56.51 & 90.37 & 37.40 & 40.33 & 56.15 \\
            & IGSP (ours)
                & 36.63 & 81.43 & 16.92 & 18.81 & \textbf{\underline{38.45}}
                & 61.25 & 90.21 & 48.65 & 49.89 & \textbf{\underline{62.50}} \\
            \midrule
            \multirow{10}{*}{SparseGPT}
            & C4
                & 60.74 & 84.35 & 0.41 & 14.47 & 39.99
                & 75.77 & 92.56 & 6.25 & 25.84 & 50.11 \\
            & Wikipedia
                & 56.36 & 81.97 & 7.08 & 18.64 & 41.01
                & 71.41 & 91.44 & 12.26 & 28.61 & 50.93 \\
            & GSM8K
                & 50.82 & 84.59 & 11.07 & 49.06 & 48.88
                & 63.50 & 91.19 & 24.76 & 68.99 & 62.11 \\
            & MetaMath
                & 46.20 & 81.83 & 19.87 & 52.16 & 50.02
                & 54.24 & 88.76 & 39.15 & 79.20 & 65.34 \\
            \cmidrule(lr){2-12}
            & Math-heavy
                & 59.64 & 85.69 & 23.45 & 45.08 & 53.46
                & 78.24 & 90.39 & 35.94 & 73.32 & 69.47 \\
            & Code-heavy
                & 58.06 & 84.00 & 59.77 & 21.76 & 55.90
                & 71.96 & 90.81 & 66.76 & 42.67 & 68.05 \\
            & Balanced
                & 59.69 & 85.32 & 52.34 & 37.76 & \textbf{58.78}
                & 72.97 & 91.45 & 62.92 & 68.65 & 74.00 \\
            \cmidrule(lr){2-12}
            & Self-Cal
                & 57.24 & 84.02 & 26.41 & 19.01 & 46.67
                & 70.63 & 92.94 & 51.36 & 44.36 & 64.82 \\
            & SGS
                & 56.22 & 85.32 & 19.04 & 16.59 & 44.29
                & 70.80 & 91.57 & 50.49 & 46.78 & 64.91 \\
            & IGSP (ours)
                & 56.09 & 85.51 & 47.34 & 39.36 & \underline{57.08}
                & 71.28 & 93.76 & 64.12 & 74.01 & \textbf{\underline{75.79}} \\
            \bottomrule
    \end{tabular}
    \caption{Capability retention at 60\% unstructured sparsity on the two main models. Magnitude uses no calibration. Standard deviations for LLaMA-3.1-8B are in the supplementary material. Boldface marks the best $S^{\mathrm{tot}}$ in each model--pruner block; underlined totals are IGSP. On Qwen2.5-7B, IGSP leads both pruners. On LLaMA-3.1-8B it leads Wanda and trails Balanced by $1.70$ under SparseGPT. Under SparseGPT on LLaMA-3.1-8B, MetaMath is the strongest of all 15 evaluated single sources by $S^{\mathrm{tot}}$ (Appendix~\ref{app:oit_profiles}); Balanced exceeds it by $8.8$ points. LLaMA-3.1-70B is in Appendix~\ref{app:llama70}.}
    \label{tab:main_results}
\end{table*}

\paragraph{Calibration data.}
Except for Magnitude, every calibration strategy uses 128 sequences of 2048 tokens~\cite{jaiswalcompressing}. The same sequences are applied at every layer, and sparsity is an unstructured ratio applied uniformly to the pruned weights. We compare three families. \emph{Single-source}: 15 public corpora spanning general text, instruction, commonsense, code, and math. \emph{Multi-source}: \textbf{Balanced} draws uniformly $1{:}1{:}1{:}1$ from the four capability pools; \textbf{Math-heavy} and \textbf{Code-heavy} reweight those pools. \emph{Self-generated}: \textbf{Self-Cal}~\cite{williams2025self}, \textbf{SGS}~\cite{ji2025beware}, and \textbf{IGSP} (capability-stratified prompts, per-dimension perplexity bands, and greedy 4-gram aggregation minimisation; \S\ref{sec:igsp-main}), with generation settings fixed across the three methods. Where a corpus also appears in evaluation, calibration uses its training split only. Corpus list, split rules, and configuration identifiers are in Appendix~\ref{app:experimental_setup}.

\paragraph{Evaluation.} Using lm-evaluation-harness~\cite{eval-harness}, we group nine benchmark configurations into four dimensions: \textsc{General} (LAMBADA-Std and LAMBADA-OpenAI~\cite{paperno2016lambada}, TriviaQA~\cite{joshi2017triviaqa}), \textsc{Commonsense} (PIQA~\cite{bisk2020piqa}, ARC-Easy~\cite{clark2018think}), \textsc{Code} (HumanEval~\cite{chen2021codex}, MBPP~\cite{austin2021program}), and \textsc{Math} (GSM8K~\cite{cobbe2021training}, Minerva~\cite{lewkowycz2022solving}). The two LAMBADA variants enter the General average as separate tasks. General and Commonsense use zero-shot evaluation; Code and Math use 3-shot evaluation. Decoding is greedy ($T{=}0$), and we report $S_{m,d}$ and its uniform average $S_m^{\text{total}}$. All reported results are means over eight random seeds. A held-out suite shares no task with calibration construction or with this main suite; its task list and scores are in \S\ref{sec:heldout} and Appendix~\ref{app:heldout}.

\section{Results}
\label{sec:results}
% v3 changes vs v2:
%  * §5.1: positioning sentence for cw4b ("limitations of averaged
%    evaluation are not new in general; our increment is a quantitative,
%    opposite-sign, sparsity-amplified characterisation in high-sparsity
%    pruning") + NEW "genuine cancellation" paragraph (plan §5).
%  * §5.2: rewritten per E2 (BH-FDR over 48 cells + 15x LOSO). Tiered
%    claims replace "robust predictors"; wording follows E2_RESULTS.md §3.
%  * §5.3/§5.4: display names unified (Balanced); coverage-vs-selection
%    reports the self-generation ablation in Table tab:igsp_ablation.

\begin{table*}[t]
    \centering
    \small
    \begin{tabular}{@{}lcccc@{\qquad}lcccc@{}}
        \toprule
        \multicolumn{5}{c}{Real-data calibration} & \multicolumn{5}{c}{Self-generation} \\
        \cmidrule(lr){1-5}\cmidrule(lr){6-10}
        Strategy & 30\% & 40\% & 50\% & 60\% & Strategy & 30\% & 40\% & 50\% & 60\% \\
        \midrule
        \multicolumn{5}{@{}l}{\emph{Single-source}} & \multicolumn{5}{l@{}}{\emph{Methods and ablations}} \\
        C4 (default) & 93.28 & 81.72 & 62.08 & 39.99 & Self-Cal & 94.50 & 84.50 & 66.75 & 46.67 \\
        Wikipedia & 93.49 & 82.84 & 63.59 & 41.01 & SGS & 94.30 & 84.10 & 67.18 & 44.29 \\
        GSM8K & 95.67 & 87.08 & 71.45 & 48.88 & \textsc{--no-multi} & 93.85 & 83.40 & 66.20 & 44.80 \\
        MetaMath & 97.83 & 90.30 & 75.12 & 50.02 & \textsc{--no-band} & 94.70 & 86.95 & 69.85 & 50.35 \\
        & & & & & \textsc{--strat.-rand.} & 94.90 & 85.60 & 71.45 & 49.20 \\
        \addlinespace
        \multicolumn{5}{@{}l}{\emph{Multi-source}} & \textsc{--no-div.} & 94.58 & 87.24 & 72.06 & 51.47 \\
        Math-heavy & 96.86 & 89.32 & 76.22 & 53.46 & IGSP (full) & \textbf{95.20} & \textbf{89.50} & \textbf{75.32} & \textbf{57.08} \\
        Code-heavy & 95.80 & 87.31 & 73.48 & 55.90 & & & & & \\
        Balanced & \textbf{97.90} & \textbf{91.34} & \textbf{79.25} & \textbf{58.78} & & & & & \\
        \bottomrule
    \end{tabular}
    \caption{SparseGPT total retention on LLaMA-3.1-8B. The left panel is the real-data comparison: Balanced leads at every sparsity, and its margin over C4 grows from $+4.62$ to $+18.79$. The right panel is the IGSP ablation: stratification, perplexity bands, and diversity each remove a smaller share than the full method, and IGSP leads the self-generation panel at every sparsity. Standard deviations are in the supplementary material.}
    \label{tab:igsp_ablation}
\end{table*}

%--------------------------------------------------------
\subsection{Single-axis Evaluation Hides Calibration Impact}
\label{sec:single_axis}

\paragraph{RQ1.} Does a single averaged score adequately capture the effects of calibration data?

\citet{williams2024impact} report that calibration choices have only modest impact on averaged accuracy across general-domain commonsense-reasoning benchmarks (PIQA, ARC, HellaSwag, BoolQ, and WinoGrande). We reproduce this impression at SparseGPT 60\%: averaged accuracy on PIQA and ARC-Easy, the commonsense subset of our evaluation suite, spans only $2.85$ points across four calibration strategies (Figure~\ref{fig:hook}b), superficially supporting the conclusion that calibration matters little.

This $2.85$-point span hides effects that are opposite in sign across capabilities and that grow with sparsity. At SparseGPT 60\%, MetaMath calibration yields $S_{m,\text{Math}}{=}52.2$, compared with $14.5$ for C4 ($+37.7$), whereas C4 yields $S_{m,\text{General}}{=}60.7$, compared with $46.2$ for MetaMath ($-14.5$). Replacing C4 with Balanced also raises $S_{m,\text{Code}}$ by $51.9$ points (Figure~\ref{fig:hook}a). The commonsense-only aggregate conceals these opposite-direction changes; capability-decomposed evaluation makes them visible.

That averaged evaluation can obscure per-task effects is known in general~\cite{he2025preserving,williams2024impact}. Two masking mechanisms should be distinguished. The first is \emph{incomplete coverage}: an average computed over a narrow benchmark subset (e.g., commonsense only) simply does not observe other dimensions (Figure~\ref{fig:hook}). The second is \emph{genuine cancellation}: under Wanda at 60\% sparsity, C4, Wikipedia, GSM8K, and MetaMath differ by at most $0.92$ points in four-dimension Total retention ($34.01$--$34.93$), yet span $7.2$ points on General, $7.1$ on Code, and $5.7$ on Math; the aggregate is nearly constant while the capability profile is not. Both mechanisms show that calibration conclusions drawn from aggregated scores do not transfer to capability-level conclusions.

\textbf{Takeaway.} Averaged scores can hide opposite-signed capability changes.

%--------------------------------------------------------
\subsection{An Opposite-sign Trade-off Across Dimensions}
\label{sec:correlation}

\paragraph{RQ2.} Which data properties are associated with capability retention?

We correlate six non-volume information metrics with four retention scores across $n{=}15$ sources; Volume is fixed by the calibration budget. Full profiles and statistics are in the supplementary material. We treat these $48$ tests ($6$ metrics $\times$ $4$ dimensions $\times$ $2$ pruners) as \emph{diagnostic} and correct for multiple comparisons (Benjamini--Hochberg, $\alpha{=}0.05$), complementing each test with a leave-one-source-out (LOSO) stability check over all $15$ subsets. Figure~\ref{fig:correlation_heatmap} shows the full matrix; three patterns emerge, which we report at different confidence tiers.

\textit{Finding 1.} Aggregation is negatively associated with General retention, and this is the only cell that survives FDR. Under Wanda, $\rho{=}{-}0.793$ ($p{=}4{\times}10^{-4}$) remains significant in all $15/15$ LOSO subsets (range $[{-}0.82,{-}0.75]$). Under SparseGPT the same direction holds nominally ($\rho{=}{-}0.58$). High 4-gram redundancy is thus reliably linked to poorer General retention.

\textit{Finding 2.} Distortion flips sign across capabilities (exploratory). Calibration perplexity correlates positively with General retention (Wanda: $\rho{=}{+}0.711$; SparseGPT: $\rho{=}{+}0.554$) and negatively with Math/Code retention (Wanda--Math: $\rho{=}{-}0.525$; SparseGPT--Code: $\rho{=}{-}0.593$). These cells are nominally significant and sign-stable under LOSO, but they do not survive FDR correction. The pattern is consistent with higher-perplexity web text favouring General and lower-perplexity structured data favouring Math or Code. Each evaluated single source therefore leaves at least one capability poorly preserved, which motivates complementary mixtures.

\textit{Finding 3.} Code associations depend on the pruner (exploratory). No metric is nominally significant for Code under Wanda, whereas Distortion reaches $\rho{=}{-}0.593$ under SparseGPT. The mechanistic account is in \S\ref{sec:wanda_saturation}.

These correlations motivate the per-dimension perplexity constraints that IGSP operationalizes; the controlled experiments in \S\ref{sec:multi_source}--\S\ref{sec:igsp_automation} test that design directly.

\textbf{Takeaway.} One correction-surviving signal (Aggregation$\to$General) and an exploratory sign flip (Distortion) are enough to reject a universal single source.

%--------------------------------------------------------
\subsection{Multi-source Mixing Mitigates the Trade-off}
\label{sec:multi_source}

\paragraph{RQ3.} Does capability-balanced mixing mitigate the trade-off across sparsity and scale?

\paragraph{Single-source calibration is structurally insufficient.}
At SparseGPT 60\% on LLaMA-3.1-8B, C4 retains $60.74$ on General but only $0.41$ on Code and $14.47$ on Math; MetaMath reaches $19.87$ on Code and $52.16$ on Math but falls to $46.20$ on General (Table~\ref{tab:main_results}). The same split appears on Qwen2.5-7B: C4 retains $75.77$ on General and $6.25$ on Code, whereas MetaMath retains $79.20$ on Math and $54.24$ on General. Wanda exhibits the same trade-off, and none of the evaluated single sources covers all four capabilities.

\paragraph{Multi-source mixing provides the largest gains at high sparsity.}
On both main models, Balanced's margin over C4 grows with sparsity. On LLaMA-3.1-8B it rises from $+4.62$ at 30\% to $+18.79$ at 60\% (Table~\ref{tab:igsp_ablation}); on Qwen2.5-7B it rises from $+3.10$ at 30\% to $+23.89$ at 60\% (Appendix Table~\ref{tab:qwen_sparsity}). IGSP stays close to Balanced on LLaMA-3.1-8B and leads it at every sparsity from 30\% to 60\% on Qwen2.5-7B, reaching $75.79$ against Balanced $74.00$, MetaMath $65.34$, and C4 $50.11$ at SparseGPT 60\% (Table~\ref{tab:cross_family}). At Wanda 60\%, IGSP is best on both models ($38.45$ and $62.50$).

\paragraph{Broad coverage matters more than targeted reweighting.}
At SparseGPT 60\% on LLaMA-3.1-8B, the Math- and Code-heavy variants trail Balanced ($53.46/55.90$ vs.\ $58.78$). On Qwen2.5-7B the same comparison is $69.47/68.05$ vs.\ $74.00$. Supplementary leave-one-out and ratio-sensitivity analyses further show that every pool contributes and nearby allocations remain within $1.50$ points, favouring broad coverage over precise weighting.

\paragraph{The ordering persists at 70B.}
At LLaMA-3.1-70B and 60\% sparsity, Balanced reaches $75.15$ under SparseGPT and $66.63$ under Wanda, exceeding the strongest displayed single sources by $+5.82$ and $+2.24$; IGSP reaches $74.78$ under SparseGPT, $0.37$ below Balanced and $+7.23$ over Self-Cal (Appendix~\ref{app:llama70}).

\begin{table}[htbp]
    \centering
    \small
    \setlength{\tabcolsep}{3.8pt}
    \begin{tabular}{@{}lcccc@{}}
        \toprule
        & \multicolumn{2}{c}{Main suite $S^{\mathrm{tot}}$} & \multicolumn{2}{c}{Held-out $S^{\mathrm{tot}}$} \\
        \cmidrule(lr){2-3} \cmidrule(lr){4-5}
        Calibration & LLaMA-8B & Qwen-7B & LLaMA-8B & Qwen-7B \\
        \midrule
        C4        & 39.99 & 50.11 & 44.37 & 56.11 \\
        MetaMath  & 50.02 & 65.34 & 56.56 & 64.46 \\
        Balanced  & \textbf{58.78} & 74.00 & \textbf{64.40} & 72.34 \\
        Self-Cal  & 46.67 & 64.82 & 52.45 & 66.84 \\
        SGS       & 44.29 & 64.91 & 52.47 & 67.88 \\
        IGSP (ours) & \underline{57.08} & \textbf{\underline{75.79}} & \underline{61.42} & \textbf{\underline{74.00}} \\
        \bottomrule
    \end{tabular}
    \caption{SparseGPT total retention at 60\% unstructured sparsity. IGSP stays within $1.70$ of Balanced on LLaMA-3.1-8B and leads on Qwen2.5-7B for both the main suite and held-out benchmarks. Boldface marks the best entry in each column; underlined entries are IGSP. LLaMA-3.1-70B is in Appendix~\ref{app:llama70}.}
    \label{tab:cross_family}
\end{table}

\textbf{Takeaway.} Coverage gains grow with sparsity on both main models and persist at 70B.

%--------------------------------------------------------
\subsection{IGSP: Multi-source Construction without Aligned Corpora}
\label{sec:igsp_automation}

\paragraph{RQ4.} Can self-generated data realise multi-source calibration, and which IGSP components matter?

Balanced presupposes curated corpora partitioned by capability. When such calibration sources are unavailable, IGSP uses the evaluation taxonomy and base model to construct capability-stratified candidate pools and select the final calibration set.

\paragraph{Ablation matrix.}
The self-generation panel of Table~\ref{tab:igsp_ablation} compares IGSP with Self-Cal, SGS, three leave-one-component ablations, and a stratified-random baseline, all under a fixed 128-sequence budget. The three ``\textsc{--no}-X'' rows remove capability stratification, per-dimension perplexity bands, or greedy 4-gram aggregation minimisation. Per-dimension numbers for IGSP, Self-Cal, and SGS at SparseGPT 60\% are in Table~\ref{tab:main_results}.

\paragraph{IGSP performs best among the evaluated self-generation methods.}
At SparseGPT 60\%, IGSP reaches $S_m^{\text{total}}{=}57.08$, outperforming Self-Cal by $+10.41$ and SGS by $+12.79$; at 50\% it reaches $75.32$, ahead of Self-Cal by $+8.57$ and SGS by $+8.14$. Removing capability stratification, perplexity bands, or diversity lowers Total by $12.28$, $6.73$, and $5.61$ points at 60\%, and by $9.12$, $5.47$, and $3.26$ points at 50\% (Table~\ref{tab:igsp_ablation}). Capability stratification is the largest contributor at both sparsities; perplexity bands and diversity each remove a smaller share.

\paragraph{Coverage vs.\ selection.}
To separate capability coverage from information-guided selection, we evaluate a stratified-random baseline that keeps IGSP's uniform per-dimension budget but replaces both PPL-band gating and greedy low-redundancy selection with uniform random draws from each pool. At SparseGPT 60\%, stratified-random reaches $49.20$, exceeding Self-Cal ($46.67$) by $2.53$ and trailing IGSP ($57.08$) by $7.88$. At 50\% it reaches $71.45$, $4.70$ above Self-Cal ($66.75$) and $3.87$ below IGSP ($75.32$). Uniform coverage alone therefore exceeds prior self-generation methods, and information-guided selection supplies a further margin that is larger at 60\%. Section~\ref{sec:heldout} shows that the same ordering holds on unseen tasks.

\paragraph{A residual gap to Balanced.}
At SparseGPT 60\%, IGSP trails Balanced by $1.70$ points (Table~\ref{tab:igsp_per_dim_gap}). The gap is largest on Code ($+5.00$); General differs by $+3.60$, Commonsense is matched ($-0.19$), and Math favors IGSP ($-1.60$). Perplexity selection therefore recovers nearly the same Total as Balanced, with the residual concentrated in Code.

\paragraph{Decision rule.}
Balanced is the strongest real-data calibration on both main models. At 70B it leads IGSP by $0.37$ under SparseGPT (Appendix~\ref{app:llama70}). Use IGSP when those corpora are unavailable: on LLaMA-3.1-8B it improves over Self-Cal and SGS by $+10.41$ and $+12.79$ at SparseGPT 60\%, and by $+8.57$ and $+8.14$ at 50\%, while trailing Balanced by $1.70$. On Qwen2.5-7B, IGSP also exceeds Balanced on the main suite and on held-out benchmarks.

\textbf{Takeaway.} Stratification is the largest IGSP component.

%--------------------------------------------------------
\subsection{When Calibration Matters: Wanda vs.\ SparseGPT}
\label{sec:wanda_saturation}

\paragraph{RQ5.} How does the sensitivity to calibration design differ between Wanda and SparseGPT at high sparsity?

A consistent observation across our experiments is that calibration design has substantially different impact under the two pruners at 60\% sparsity (Figure~\ref{fig:wanda_saturation}). Under SparseGPT, the spread between the worst (C4, $39.99$) and the best (Balanced, $58.78$) is 18.79 points. Under Wanda, the same set spans $34.01$--$38.45$, a 4.44-point window that is more than four times narrower. Within the self-generation track, Self-Cal, SGS, and IGSP span $36.02$--$38.45$ (2.43 points), with IGSP providing the strongest Wanda result. On Qwen2.5-7B the direction is the same and the Wanda window is again the narrower of the two, but both windows are wider than on LLaMA-3.1-8B: Balanced and C4 differ by $23.89$ points under SparseGPT and by $13.93$ under Wanda (Table~\ref{tab:main_results}).

\paragraph{Mechanistic explanation.}
SparseGPT folds calibration into a layer-wise OBS-style update, so the samples both score weights and set the compensation applied to the survivors. Wanda uses the same samples only to compute activation magnitudes, and leaves surviving weights unchanged. This extra update pathway is consistent with the wider SparseGPT spread; the spread itself does not establish the mechanism. The update formulas are in Appendix~\ref{app:mechanism}.

\paragraph{Takeaway.}
Calibration claims should be reported across pruners and sparsity levels.

%--------------------------------------------------------
\subsection{Transfer to Held-out Benchmarks}
\label{sec:heldout}
% 2026-10-13: compressed. Task list, per-dimension detail, and the
% main-vs-held-out comparison live in Appendix app:heldout.
% Optimized IGSP: LLaMA held-out 61.42 (trails Balanced 64.40 by 2.98);
% Qwen held-out 74.00 (leads Balanced 72.34).

\paragraph{RQ6.} Do the calibration conclusions transfer to capability benchmarks that were never used during construction?

We re-evaluate every calibration strategy at SparseGPT 60\% on a suite that shares no task with calibration construction or the main evaluation (Table~\ref{tab:cross_family}). The benchmark list and per-dimension scores are in Appendix~\ref{app:heldout}. On LLaMA-3.1-8B the three main-suite findings transfer. Single sources still trade capabilities: GSM8K reaches $84.5$ on Math and $7.0$ on Code. Balanced leads at $64.40$, $+7.8$ over MetaMath. IGSP reaches $61.42$, $+5.7$ over stratified-random and $+9.0$ over Self-Cal and SGS, and trails Balanced by $2.98$. On Qwen2.5-7B, IGSP leads at $74.00$, above Balanced ($72.34$); GSM8K again keeps Math ($88.63$) and loses Code ($27.15$).

\textbf{Takeaway.} The calibration conclusions transfer to tasks never used during construction.

\section{Conclusion}
% 2026-10-13: compressed to one paragraph for the 8-page limit.
% Numbers follow the optimized IGSP rows (LLaMA-8B 57.08, 70B 74.78,
% Qwen main 75.79, Qwen held-out 74.00).

Calibration for high-sparsity pruning should be chosen for capability coverage. Balanced tests that principle when aligned corpora exist; IGSP approximates it when they do not. On LLaMA-3.1-8B the SparseGPT margin of Balanced over C4 grows from $4.6$ points at 30\% sparsity to $18.8$ at 60\% (Table~\ref{tab:igsp_ablation}), and IGSP leads Balanced on Qwen2.5-7B for both the main suite and held-out benchmarks (Table~\ref{tab:cross_family}). Those claims should be checked across pruners: at 60\% sparsity on LLaMA-3.1-8B the calibration spread is $18.8$ points under SparseGPT and $4.4$ under Wanda.

\section*{Limitations}
% COLING (via ARR) requires a standalone Limitations section; it does not
% count toward the 8-page limit.

Our study focuses on three open-source decoder-only LLMs (LLaMA-3.1-8B/70B and Qwen2.5-7B) and widely used post-training pruners (Wanda, SparseGPT, and calibration-free Magnitude) under unstructured sparsity up to 60\% and a fixed calibration budget of 128 sequences of 2048 tokens. While this setting covers common practical deployments, it does not exhaust the full space of architectures (e.g., encoder--decoder or MoE), compression schemes (e.g., structured N:M sparsity, joint pruning--quantization, or post-compression fine-tuning), or denser calibration budgets. In addition, our evaluation centers on English text and four capability dimensions (General, Commonsense, Code, and Math); we leave multilingual, multimodal, and safety-related behaviors for future work. Finally, IGSP and the information profile that motivates it involve concrete design choices (e.g., the capability taxonomy, the source-corpus perplexity bands, and 4-gram aggregation) that are reasonable but not unique. We view these aspects as natural directions to generalize and stress-test the framework, rather than fundamental constraints on the applicability of capability-balanced or information-guided calibration to LLM pruning.

\bibliography{custom}

\appendix
% -----------------------
% Appendix (\appendix is declared in main.tex)
% -----------------------

\section{Methodological Details}
\label{app:method}

\subsection{Full Definition of the Multi-Dimensional Evaluation Framework}
\label{app:multidim-eval}

Let $\mathcal{M}$ be the set of models under consideration, including the unpruned base model $m_0$ and pruned models $m \in \mathcal{M}\setminus\{m_0\}$. For each model $m$, task $t \in \mathcal{T}$, and random seed $s \in \mathcal{S}$, let
\[
y_{m,s,t} \in [0,1]
\]
denote the raw task score from one independent evaluation run (e.g., accuracy or Pass@1/Pass@k). We first average across seeds for a fixed model--task pair:
\begin{equation}
\bar{y}_{m,t} = \frac{1}{|\mathcal{S}|} \sum_{s \in \mathcal{S}} y_{m,s,t}.
\end{equation}

Using the unpruned base model $m_0$ as the reference, we define the task-level relative retention as
\begin{equation}
\tilde{y}_{m,t} = \frac{\bar{y}_{m,t}}{\bar{y}_{m_0,t} + \varepsilon},
\end{equation}
where $\varepsilon$ is a small constant for numerical stability (e.g., $10^{-6}$). At the capability-dimension level, we define
\begin{equation}
S_{m,d} = \frac{1}{|\mathcal{T}_d|} \sum_{t \in \mathcal{T}_d} \tilde{y}_{m,t},
\end{equation}
and the overall multi-dimensional score as
\begin{equation}
S^{\text{total}}_{m} = \frac{1}{|\mathcal{D}|} \sum_{d \in \mathcal{D}} S_{m,d}.
\end{equation}

When $\tilde{y}_{m,t} \approx 1$, the pruned model $m$ can be regarded as largely preserving the capability of $m_0$ on task $t$. This normalization makes comparisons across tasks, sparsity levels, pruning methods, and model sizes consistent.

\subsection{Precise Definitions of the OIT-Based Information Metrics}
\label{app:oit-metrics}

Let the calibration set be $C = \{x_i\}_{i=1}^N$ with text field $\mathrm{text}_i$. Let $\mathrm{Tok}(\cdot)$ be the tokenizer of the target model.

\paragraph{Volume.}
$\mathrm{total\_tokens}(C) = \sum_{i=1}^{N} |\mathrm{Tok}(\mathrm{text}_i)|$. In practice, we treat total tokens as a hard constraint (token budget) when comparing calibration strategies.

\paragraph{Scope.}
Let $\mathcal{D}_{\mathrm{cov}} = \bigcup_{i=1}^{N} \mathcal{D}_i$ be the set of capability dimensions covered by $C$, where $\mathcal{D}_i \subseteq \mathcal{D} = \{\text{General}, \text{Commonsense}, \text{Code}, \text{Math}\}$. Then $\mathrm{Scope}(C) = |\mathcal{D}_{\mathrm{cov}}| / |\mathcal{D}|$.

\paragraph{Variety.}
Variety measures genre richness. For genre set $\mathcal{G}$, let $k = |\{ g \in \mathcal{G} : n_g > 0\}|$ be the number of observed genres in $C$, and $k_{\max} = |\mathcal{G}^{\mathrm{global}}|$. Then $\mathrm{Variety}(C) = k / k_{\max}$ (0 if $k_{\max}=0$).

\paragraph{Granularity.}
Let $\Lambda(C) = \{(d, g) \mid \exists\, x_i \in C,\ d \in \mathcal{D}_i,\ g_i = g\}$ be the set of non-empty (dimension, genre) cells, with $K = |\Lambda(C)|$. Then $\mathrm{Granularity}(C) = 1/K$ (with $1$ if $K=0$). A calibration set occupying more distinct cells has a \emph{smaller} Granularity value.

\paragraph{Distortion.}
For each sequence $(w_{i,1},\dots,w_{i,L_i})$ in $C$, compute the average negative log-likelihood $\ell_i = -\frac{1}{L_i} \sum_{t=1}^{L_i} \log p_M(w_{i,t} \mid w_{i,<t})$. Then $\mathrm{Distortion}(C) = \exp(\frac{1}{N'} \sum_{i=1}^{N'} \ell_i)$, i.e., perplexity under the base model $M$. Larger values indicate calibration data more ``surprising'' relative to the model's distribution.

\paragraph{Mismatch.}
We adopt the Min-K\%++ statistic~\cite{shi2024min}. At each token position $t$, compute the Z-score $z_t = (\ell_{t,w_t} - \mu_t) / \sigma_t$ where $\mu_t, \sigma_t^2$ are the mean and variance of log-probabilities under the model's predictive distribution. For each example, we sort $\{z_t\}$ and take the bottom $k\%$ (hardest tokens); $\mathrm{MinK}^{++}_k(x_i)$ is their mean. $\mathrm{Mismatch}(C)$ is the average over all examples with valid scores.

\paragraph{Aggregation.}
For fixed $n$-gram length $n=4$, let $\mathcal{G}_n$ be the multiset of all contiguous $n$-grams from whitespace-tokenized text, and $\mathcal{G}^{\mathrm{uniq}}_n$ its set of unique elements. Then $\mathrm{Aggregation}(C) = 1 - |\mathcal{G}^{\mathrm{uniq}}_n| / |\mathcal{G}_n|$. Higher values indicate more repeated $n$-grams (stronger local redundancy).

\subsection{IGSP Calibration Construction}
\label{app:igsp}

Algorithm~\ref{alg:igsp_v2} constructs the IGSP calibration set. The capability taxonomy $\mathcal{D}$ follows the evaluation protocol, the allocation $\alpha_d$ is uniform by default, and the PPL bands $[\tau_d^{\text{low}}, \tau_d^{\text{high}}]$ are obtained from a one-time PPL profile of the \emph{published source corpora} assigned to each dimension under the unpruned model (e.g., GSM8K-train and MetaMath for Math), not from the evaluation tasks. All three are fixed before pruning and are never selected using post-pruning retention scores.

\begin{algorithm}[t]
\caption{IGSP Calibration Construction}
\label{alg:igsp_v2}
\begin{algorithmic}[1]
\REQUIRE Base model $M$, capability taxonomy $\mathcal{D} = \{d_1, \dots, d_{|\mathcal{D}|}\}$, token budget $B$, dimension weights $\{\alpha_d\}_{d \in \mathcal{D}}$ (default uniform), PPL bands $\{[\tau_d^{\text{low}}, \tau_d^{\text{high}}]\}_{d \in \mathcal{D}}$
\ENSURE Calibration set $C^*$
\STATE \textbf{Phase 1: Capability-stratified pool construction}
\FOR{each dimension $d \in \mathcal{D}$}
    \STATE Construct candidate pool $P_d$ by either:
    \STATE \quad (a) selecting real corpora aligned with $d$, or
    \STATE \quad (b) prompting $M$ with $d$-specific templates to generate synthetic data
\ENDFOR
\STATE $P \leftarrow \bigcup_{d \in \mathcal{D}} P_d$
\STATE
\STATE \textbf{Phase 2: Aggregation-constrained, Distortion-balanced selection}
\STATE Initialize $C^* \leftarrow \emptyset$
\FOR{each dimension $d \in \mathcal{D}$}
    \STATE $b_d \leftarrow \alpha_d \cdot B$ \quad (sample budget for dimension $d$)
    \STATE Filter $P_d^{\text{valid}} \leftarrow \{x \in P_d : \mathrm{PPL}(\{x\}) \in [\tau_d^{\text{low}}, \tau_d^{\text{high}}]\}$
    \STATE Select $b_d$ samples from $P_d^{\text{valid}}$ via greedy $n$-gram diversity maximization:
    \STATE \quad Greedily add samples that maximize $|\bigcup \mathrm{ngrams}(C^* \cup \{x\})| / \sum |\mathrm{ngrams}(C^* \cup \{x\})|$
    \STATE Add selected samples to $C^*$
\ENDFOR
\STATE \textbf{return} $C^*$
\end{algorithmic}
\end{algorithm}

The PPL bands $[\tau_d^{\text{low}}, \tau_d^{\text{high}}]$ are estimated once from the empirical PPL distributions of the published source corpora assigned to dimension $d$ under the unpruned model. They are then held fixed across candidate pools, pruning methods, sparsity levels, and random seeds, with no iterative optimization against post-pruning performance; no evaluation-task input or post-pruning score participates in setting them. The source corpora and numerical values are reported in Appendix~\ref{app:igsp_ppl_bands}.

%---------------------------------------------------------------------

\section{Calibration Data: Expanded Profiles and Baselines}

\subsection{OIT Information Profiles for All 15 Calibration Datasets}
\label{app:oit_profiles}

\begin{table*}[h]
    \centering
    \caption{Full OIT information profiles for all $n=15$ calibration datasets (LLaMA-3.1-8B base model). All measurements are on 128-sample calibration batches with fixed token budget. Datasets are ordered by Distortion (PPL). \textbf{Bold} = extreme values per column.}
    \label{tab:oit_profiles_15}
    \small
    \setlength{\tabcolsep}{3pt}
    \begin{tabular}{lcccccc}
        \toprule
        Dataset & Scope & Variety & Granularity & Aggregation & Distortion (PPL) & Mismatch \\
        \midrule
        MetaMath       & 0.75 & 0.0000 & 0.3333 & 0.3000 & \textbf{2.2333} & \textbf{-1.8783} \\
        Code-Alpaca    & 0.50 & 0.1161 & 0.5000 & 0.4447 & 2.2816 & -2.2361 \\
        GSM8K          & 0.75 & 0.1667 & 0.3333 & 0.1481 & 3.0635 & -2.2060 \\
        Alpaca         & 1.00 & 0.6277 & 0.0909 & 0.3419 & 3.8959 & -2.3909 \\
        TinyStories    & 0.50 & 0.0659 & 0.5000 & 0.2323 & 3.9883 & -2.3133 \\
        MBPP           & 0.50 & 0.5000 & 0.2500 & \textbf{0.9161} & 2.6473 & -2.2339 \\
        Dolly          & 0.75 & 0.4019 & 0.1000 & 0.2019 & 5.2180 & -2.3057 \\
        WinoGrande     & 0.50 & 0.5151 & 0.2000 & 0.7284 & 5.5046 & -2.2048 \\
        HellaSwag      & 0.75 & 0.3333 & 0.2000 & 0.1196 & 6.4061 & -2.2051 \\
        Wikipedia      & 1.00 & 0.6667 & 0.1000 & 0.0314 & 6.4145 & -2.3382 \\
        WikiText2      & 0.50 & 0.5000 & 0.1667 & 0.0727 & 6.5678 & -2.1989 \\
        arXiv          & 1.00 & 0.1147 & 0.1667 & 0.1176 & 7.8356 & -2.3131 \\
        C4             & 0.25 & 0.1667 & 1.0000 & 0.0594 & 8.3216 & -2.1317 \\
        OpenBookQA     & 0.50 & 0.7304 & 0.1667 & 0.6514 & \textbf{12.9508} & -2.2092 \\
        DCLM           & 1.00 & \textbf{1.0000} & 0.0625 & \textbf{0.0146} & \textbf{11.1965} & -2.3432 \\
        \bottomrule
    \end{tabular}
\end{table*}

Key observations from Table~\ref{tab:oit_profiles_15}:
\begin{itemize}
    \item \textbf{Distortion spans an order of magnitude} from 2.23 (MetaMath) to 12.95 (OpenBookQA). Task-specific data (Math, Code) occupies the low-PPL regime ($\approx 2$--$4$); general web text occupies the high-PPL regime ($\approx 6$--$13$). This large spread makes single-PPL calibration intrinsically incapable of covering the full capability space.
    \item \textbf{Aggregation highlights structural differences.} MBPP (code) has the highest redundancy (0.92) due to repeated code patterns; DCLM has the lowest (0.01) due to diverse web text. The strong negative correlation between Aggregation and General retention ($\rho=-0.793$) means that high-redundancy data systematically degrades general language capability.
    \item \textbf{Mismatch has narrow dynamic range} (only 0.51 units from $-1.88$ to $-2.39$ across all 15 datasets), explaining its failure to predict capability retention in our correlation analysis.
\end{itemize}

\subsection{Single-Source Calibration at 40\% Sparsity (Detailed)}
\label{app:single_source_40}

\begin{table}[h]
    \centering
    \caption{Single-source calibration results at 40\% unstructured sparsity, LLaMA-3.1-8B. The Total spread across 7 sources is only 2.7 points (Wanda) and 6.1 points (SparseGPT)---but per-dimension gaps exceed 19 points (Math under SparseGPT). This table supports the claim in Section~\ref{sec:single_axis} that single-axis evaluation hides calibration impact.}
    \label{tab:single_source_40}
    \small
    \setlength{\tabcolsep}{2pt}
    \begin{tabular}{llccccc}
        \toprule
        Method & Data & $S_{\text{Gen}}$ & $S_{\text{Com}}$ & $S_{\text{Code}}$ & $S_{\text{Math}}$ & $S^{\text{total}}$ \\
        \midrule
        \multirow{7}{*}{Wanda}
        & C4          & 91.61 & 96.39 & 76.81 & 64.17 & 82.25 \\
        & DCLM        & 90.88 & 96.25 & 75.80 & 63.09 & 81.51 \\
        & Wikipedia   & 90.61 & 96.60 & 81.25 & 67.06 & 83.88 \\
        & WikiText2   & 91.01 & 96.79 & 78.66 & 68.10 & 83.64 \\
        & GSM8K       & 87.56 & 95.77 & 78.38 & \textbf{72.62} & 83.58 \\
        & MBPP        & 87.49 & 96.00 & 80.56 & 67.11 & 82.79 \\
        & HellaSwag   & 90.84 & 96.40 & 73.82 & 63.56 & 81.16 \\
        \midrule
        & \textbf{Spread} & 4.05 & 1.02 & 7.43 & 9.06 & 2.72 \\
        \midrule
        \multirow{7}{*}{SparseGPT}
        & C4          & 90.90 & 97.57 & 71.96 & 67.80 & 82.06 \\
        & DCLM        & 91.40 & 97.74 & 73.44 & 68.89 & 82.87 \\
        & Wikipedia   & \textbf{92.73} & 97.62 & 72.27 & 67.60 & 82.56 \\
        & WikiText2   & 92.18 & 97.70 & 72.26 & 66.55 & 82.17 \\
        & GSM8K       & 87.24 & 97.88 & 76.27 & \textbf{87.48} & \textbf{87.22} \\
        & MBPP        & 76.53 & 96.06 & \textbf{91.39} & 67.40 & 82.85 \\
        & HellaSwag   & 89.41 & \textbf{98.40} & 76.61 & 66.29 & 82.68 \\
        \midrule
        & \textbf{Spread} & 16.20 & 2.34 & 19.43 & 21.19 & 6.15 \\
        \bottomrule
    \end{tabular}
\end{table}

Table~\ref{tab:single_source_40} provides the complete single-source results at 40\% sparsity for both Wanda and SparseGPT. Under Wanda, the 7-source Total spread is only 2.72 points, consistent with the ``calibration matters little'' finding of~\citet{williams2024impact}. But the per-dimension spread tells a different story: Math retention varies by 9.06 points (63.56--72.62), and under SparseGPT, both Math (21.19 points) and Code (19.43 points) exhibit spreads that are invisible in the Total average.

%---------------------------------------------------------------------

\section{Experimental Setup Details}
\label{app:experimental_setup}

\subsection{Calibration Data and Formats}
We construct calibration sets from 15 public corpora. General-domain text: C4~\cite{raffel2020exploring}, DCLM~\cite{Li2024DataCompLM}, Wikipedia, WikiText-2~\cite{team2024gemma}, and arXiv. Instruction: Alpaca, Dolly, and TinyStories. Commonsense: HellaSwag, OpenBookQA, and WinoGrande. Code: MBPP and Code-Alpaca. Math: GSM8K and MetaMath. Multi-source mixes partition these corpora by their primary capability. Displayed names map to configuration identifiers as Balanced${=}$\texttt{mix\_balanced} (uniform $1{:}1{:}1{:}1$), Math-heavy${=}$\texttt{mix\_4\_3\_3\_math}, and Code-heavy${=}$\texttt{mix\_4\_3\_3\_code}. Each record is a JSON object; for text corpora the primary field is \texttt{text}, and for reasoning datasets we follow canonical fields (e.g., \texttt{question}/\texttt{answer} in GSM8K). Calibration sequences for the two corpora that also appear in the evaluation suite (GSM8K, MBPP) are loaded from their \emph{train} splits (\texttt{train-00000-of-00001.parquet}); evaluation via lm-evaluation-harness uses the disjoint \emph{test} splits, so no calibration example is ever scored. MetaMath is derived from GSM8K's train questions and is likewise disjoint from the GSM8K test set. The commonsense calibration corpora are not part of the evaluation suite.

\subsection{Tasks and Taxonomy}
\label{app:tasks_taxonomy}

We group tasks into four dimensions:
\textbf{General (Gen)}: LAMBADA (Standard + OpenAI), TriviaQA;
\textbf{Commonsense (Com)}: PIQA, ARC-Easy;
\textbf{Code (Code)}: HumanEval, MBPP;
\textbf{Math (Math)}: GSM8K, Minerva Math.
Gen/Com tasks are evaluated zero-shot; Code/Math tasks are evaluated 3-shot, consistent with standard protocol. All benchmarks are in English and evaluated with greedy decoding ($T=0$, batch\_size=32, max\_length=2048, up to 1000 examples per task).

\subsection{Hardware and Reproducibility}
All experiments run on NVIDIA H800 GPUs (80GB). Each run uses 1--2 GPUs depending on model size. Key configurations are repeated over 8 random seeds \{0, 5, 10, 15, 20, 25, 30, 35\}; we report means and standard deviations where applicable. Exact data snapshots, filtering rules, and seeds are in the released code.

\subsection{Standard Deviations for Main-Paper Tables}

Table~\ref{tab:main_results_sd} reports standard deviations for the LLaMA models. The LLaMA-3.1-8B columns pair with Table~\ref{tab:main_results}; the LLaMA-3.1-70B columns pair with Table~\ref{tab:llama70}. Table~\ref{tab:igsp_ablation_sd} pairs with both panels of Table~\ref{tab:igsp_ablation}.

\begin{table*}[t]
    \centering
    \small
    \begin{tabular}{@{}llccccc@{\qquad}ccccc@{}}
        \toprule
        & & \multicolumn{5}{c}{LLaMA-3.1-8B} & \multicolumn{5}{c}{LLaMA-3.1-70B} \\
        \cmidrule(lr){3-7}\cmidrule(lr){8-12}
        Method & Calibration & $S_{\mathrm{Gen}}$ & $S_{\mathrm{Com}}$ & $S_{\mathrm{Code}}$ & $S_{\mathrm{Math}}$ & $S^{\mathrm{tot}}$
                              & $S_{\mathrm{Gen}}$ & $S_{\mathrm{Com}}$ & $S_{\mathrm{Code}}$ & $S_{\mathrm{Math}}$ & $S^{\mathrm{tot}}$ \\
        \midrule
        Magnitude & None & 0.12 & 0.05 & 0.18 & 0.20 & 0.10 & 0.20 & 0.10 & 0.42 & 0.48 & 0.22 \\
        \midrule
        \multirow{10}{*}{Wanda}
        & C4          & 0.55 & 0.22 & 0.20 & 0.50 & 0.42 & 0.48 & 0.18 & 0.42 & 0.45 & 0.35 \\
        & Wikipedia   & 0.58 & 0.22 & 0.55 & 0.55 & 0.45 & 0.50 & 0.20 & 0.48 & 0.48 & 0.40 \\
        & GSM8K       & 0.65 & 0.28 & 0.62 & 0.85 & 0.55 & 0.55 & 0.25 & 0.55 & 0.72 & 0.42 \\
        & MetaMath    & 0.62 & 0.25 & 0.58 & 0.78 & 0.50 & 0.52 & 0.22 & 0.50 & 0.68 & 0.45 \\
        \cmidrule(lr){2-12}
        & Math-heavy  & 0.48 & 0.22 & 0.55 & 0.55 & 0.40 & 0.42 & 0.20 & 0.48 & 0.48 & 0.32 \\
        & Code-heavy  & 0.50 & 0.22 & 0.85 & 0.50 & 0.45 & 0.45 & 0.20 & 0.72 & 0.45 & 0.40 \\
        & Balanced    & 0.42 & 0.20 & 0.62 & 0.55 & 0.38 & 0.38 & 0.18 & 0.55 & 0.48 & 0.30 \\
        \cmidrule(lr){2-12}
        & Self-Cal    & 0.78 & 0.30 & 0.92 & 0.85 & 0.62 & 0.65 & 0.25 & 0.78 & 0.72 & 0.55 \\
        & SGS         & 0.88 & 0.32 & 1.05 & 0.95 & 0.70 & 0.75 & 0.28 & 0.88 & 0.82 & 0.62 \\
        & IGSP (ours) & 0.68 & 0.28 & 0.78 & 0.72 & 0.61 & 0.58 & 0.22 & 0.65 & 0.62 & 0.52 \\
        \midrule
        \multirow{10}{*}{SparseGPT}
        & C4          & 0.72 & 0.30 & 0.18 & 0.65 & 0.62 & 0.60 & 0.25 & 0.45 & 0.55 & 0.50 \\
        & Wikipedia   & 0.75 & 0.32 & 0.62 & 0.78 & 0.58 & 0.62 & 0.28 & 0.55 & 0.65 & 0.48 \\
        & GSM8K       & 0.88 & 0.32 & 0.92 & 1.20 & 0.78 & 0.72 & 0.28 & 0.78 & 1.00 & 0.65 \\
        & MetaMath    & 0.92 & 0.35 & 1.10 & 1.25 & 0.82 & 0.78 & 0.30 & 0.95 & 1.05 & 0.68 \\
        \cmidrule(lr){2-12}
        & Math-heavy  & 0.65 & 0.28 & 0.92 & 1.05 & 0.65 & 0.55 & 0.25 & 0.80 & 0.88 & 0.55 \\
        & Code-heavy  & 0.72 & 0.32 & 1.40 & 0.92 & 0.88 & 0.62 & 0.28 & 1.18 & 0.78 & 0.72 \\
        & Balanced    & 0.55 & 0.25 & 0.95 & 0.95 & 0.55 & 0.48 & 0.22 & 0.82 & 0.80 & 0.45 \\
        \cmidrule(lr){2-12}
        & Self-Cal    & 1.05 & 0.40 & 1.40 & 1.15 & 0.92 & 0.88 & 0.32 & 1.18 & 0.98 & 0.78 \\
        & SGS         & 1.18 & 0.42 & 1.55 & 1.30 & 1.05 & 1.00 & 0.35 & 1.32 & 1.10 & 0.85 \\
        & IGSP (ours) & 0.92 & 0.32 & 1.10 & 0.95 & 0.68 & 0.78 & 0.28 & 0.92 & 0.82 & 0.58 \\
        \bottomrule
    \end{tabular}
    \caption{Standard deviations for capability retention at 60\% unstructured sparsity. The LLaMA-3.1-8B columns pair with Table~\ref{tab:main_results}; the LLaMA-3.1-70B columns pair with Table~\ref{tab:llama70}.}
    \label{tab:main_results_sd}
\end{table*}

\begin{table*}[t]
    \centering
    \small
    \begin{tabular}{@{}lcccc@{\qquad}lcccc@{}}
        \toprule
        \multicolumn{5}{c}{Real-data calibration} & \multicolumn{5}{c}{Self-generation} \\
        \cmidrule(lr){1-5}\cmidrule(lr){6-10}
        Strategy & 30\% & 40\% & 50\% & 60\% & Strategy & 30\% & 40\% & 50\% & 60\% \\
        \midrule
        C4 (default) & 0.22 & 0.35 & 0.48 & 0.62 & Self-Cal & 0.45 & 0.65 & 0.80 & 0.92 \\
        Wikipedia    & 0.20 & 0.32 & 0.45 & 0.58 & SGS & 0.52 & 0.72 & 0.88 & 1.05 \\
        GSM8K        & 0.28 & 0.45 & 0.58 & 0.78 & \textsc{--no-multi} & 0.50 & 0.70 & 0.85 & 1.00 \\
        MetaMath     & 0.30 & 0.48 & 0.62 & 0.82 & \textsc{--no-band} & 0.40 & 0.58 & 0.72 & 0.85 \\
        Math-heavy   & 0.25 & 0.40 & 0.50 & 0.65 & \textsc{--no-div.} & 0.35 & 0.50 & 0.62 & 0.75 \\
        Code-heavy   & 0.32 & 0.52 & 0.68 & 0.88 & IGSP (full) & 0.32 & 0.48 & 0.58 & 0.68 \\
        Balanced     & 0.20 & 0.32 & 0.42 & 0.55 & & & & & \\
        \bottomrule
    \end{tabular}
    \caption{Standard deviations for SparseGPT total retention across sparsities on LLaMA-3.1-8B. Rows pair with the two panels of Table~\ref{tab:igsp_ablation}.}
    \label{tab:igsp_ablation_sd}
\end{table*}

\subsection{Impact of Sparsity and Structure}
\label{app:sparsity_impact}

The results are shown in Table~\ref{tab:sparsity_relative}.

\begin{figure*}[htbp]
  \includegraphics[width=\textwidth]{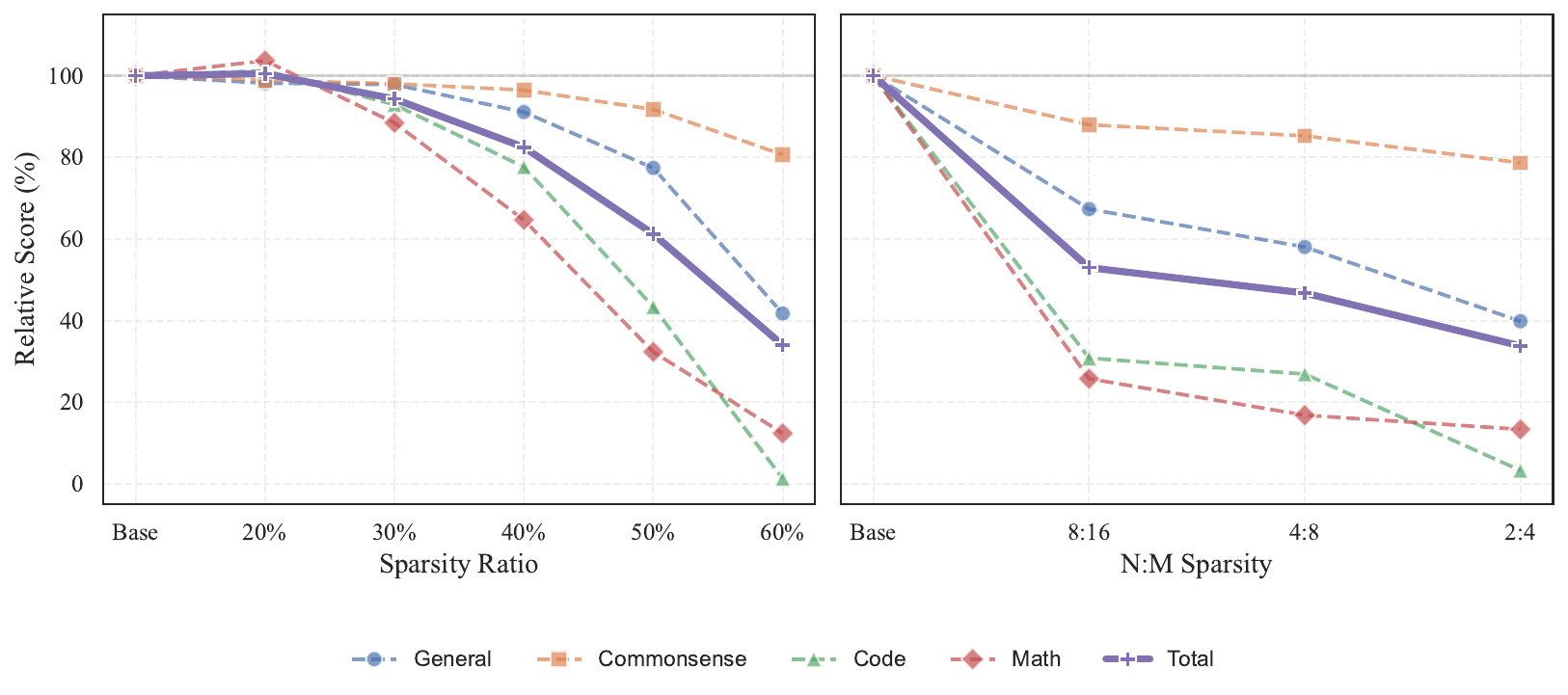}
  \caption{Impact of sparsity on model performance across different domains under Wanda with C4 calibration. Left: unstructured sparsity from 20\% to 60\%. Right: N:M semi-structured sparsity.}
  \label{fig:avg_results}
\end{figure*}

\begin{table}[htbp]
    \centering
    \caption{Relative scores ($S_{m,d}$) of Wanda pruning across sparsity configurations (C4 calibration, LLaMA-3.1-8B).}
    \label{tab:sparsity_relative}
    \small
    \setlength{\tabcolsep}{2pt}
    \begin{tabular}{lccccc}
        \toprule
        Config & $S_{m,\text{Gen}}$ & $S_{m,\text{Com}}$ & $S_{m,\text{Code}}$ & $S_{m,\text{Math}}$ & $S_{m}^{\text{total}}$ \\
        \midrule
        Base & 100.00 & 100.00 & 100.00 & 100.00 & 100.00 \\
        \midrule
        20\%  & 98.16 & 98.89 & 101.23 & 103.69 & 100.49 \\
        30\%  & 97.83 & 98.02 & 92.91 & 88.44 & 94.30 \\
        40\%  & 91.08 & 96.47 & 77.65 & 64.66 & 82.46 \\
        50\%  & 77.39 & 91.73 & 43.39 & 32.32 & 61.21 \\
        60\%  & 41.72 & 80.63 & 1.31  & 12.37 & 34.01 \\
        \midrule
        2:4   & 39.84 & 78.66 & 3.30  & 13.36 & 33.79 \\
        4:8   & 58.03 & 85.26 & 26.89 & 16.82 & 46.75 \\
        8:16  & 67.32 & 87.97 & 30.81 & 25.76 & 52.96 \\
        \bottomrule
    \end{tabular}
\end{table}

%---------------------------------------------------------------------

\section{Correlation Analysis: Extended n=15 Results}
\label{app:oit_corr_pca}

\subsection{Repeated Sampling Protocol (n=15)}
\label{app:seed_protocol}
We use 15 calibration sources (7 original + 8 supplementary, listed in Appendix~\ref{app:oit_profiles}). For each source, we sample 128 calibration instances using 8 random seeds, yielding 120 runs per (model, method) configuration. For each run, we compute (i) its OIT metric vector and (ii) the corresponding post-pruning retention scores. Seed repetition captures both within-dataset sampling variability and downstream pruning sensitivity.

\subsection{Spearman Correlation Tables (n=15)}
\label{app:oit_corr_tables}

This subsection supports the claim in \S\ref{sec:correlation}: only Aggregation--General survives FDR correction, while Distortion's sign flips across capabilities.

\begin{table*}[h]
    \centering
    \caption{Spearman $\rho$ correlations between OIT metrics and capability retention, $n{=}15$ datasets, LLaMA-3.1-8B, at 40\% sparsity, recomputed from the verified 8B raw logs. \textbf{Bold} = survives Benjamini--Hochberg FDR correction ($\alpha{=}0.05$) over all 48 cells; $^{*}$ = nominal $p<0.05$ (exploratory); \textit{italic} = $p<0.10$. LOSO = fraction of the 15 leave-one-source-out subsets in which the cell reaches $p<0.05$.}
    \label{tab:corr_n15_8b}
    \small
    \setlength{\tabcolsep}{3pt}
    \begin{tabular}{llccc|ccc}
        \toprule
        & & \multicolumn{3}{c|}{\textbf{Wanda-8B}} & \multicolumn{3}{c}{\textbf{SparseGPT-8B}} \\
        \cmidrule(lr){3-5} \cmidrule(lr){6-8}
        OIT Metric & Dim & $\rho$ & $p$-val & LOSO & $\rho$ & $p$-val & LOSO \\
        \midrule
        \multirow{4}{*}{Scope}
        & Gen  & +0.145 & 0.607 & 0\%   & +0.212 & 0.447 & 0\% \\
        & Com  & -0.034 & 0.905 & 0\%   & +0.165 & 0.556 & 0\% \\
        & Code & +0.087 & 0.759 & 0\%   & +0.308 & 0.263 & 0\% \\
        & Math & -0.064 & 0.821 & 0\%   & \textit{+0.478} & \textit{0.072} & 13\% \\
        \midrule
        \multirow{4}{*}{Variety}
        & Gen  & +0.186 & 0.507 & 0\%   & +0.556$^{*}$ & 0.031 & 67\% \\
        & Com  & +0.263 & 0.344 & 0\%   & +0.120 & 0.670 & 0\% \\
        & Code & -0.104 & 0.713 & 0\%   & -0.317 & 0.250 & 0\% \\
        & Math & -0.361 & 0.186 & 7\%   & -0.295 & 0.286 & 7\% \\
        \midrule
        \multirow{4}{*}{Granularity}
        & Gen  & -0.338 & 0.218 & 7\%   & -0.585$^{*}$ & 0.022 & 80\% \\
        & Com  & -0.380 & 0.163 & 7\%   & -0.119 & 0.674 & 0\% \\
        & Code & +0.013 & 0.965 & 0\%   & +0.000 & 1.000 & 0\% \\
        & Math & +0.295 & 0.286 & 7\%   & -0.041 & 0.884 & 0\% \\
        \midrule
        \multirow{4}{*}{Aggregation}
        & Gen  & \textbf{-0.793} & \textbf{0.0004} & \textbf{100\%} & -0.579$^{*}$ & 0.024 & 67\% \\
        & Com  & -0.125 & 0.657 & 0\%   & +0.246 & 0.376 & 7\% \\
        & Code & -0.150 & 0.594 & 0\%   & +0.189 & 0.499 & 0\% \\
        & Math & -0.014 & 0.960 & 0\%   & -0.118 & 0.676 & 0\% \\
        \midrule
        \multirow{4}{*}{Distortion (PPL)}
        & Gen  & +0.711$^{*}$ & 0.003 & 100\% & +0.554$^{*}$ & 0.032 & 67\% \\
        & Com  & +0.536$^{*}$ & 0.040 & 67\%  & -0.154 & 0.585 & 0\% \\
        & Code & -0.307 & 0.266 & 0\%   & -0.593$^{*}$ & 0.020 & 80\% \\
        & Math & -0.525$^{*}$ & 0.045 & 53\%  & -0.436 & 0.105 & 0\% \\
        \midrule
        \multirow{4}{*}{Mismatch}
        & Gen  & +0.004 & 0.990 & 0\%   & -0.161 & 0.567 & 0\% \\
        & Com  & -0.364 & 0.182 & 0\%   & +0.189 & 0.499 & 0\% \\
        & Code & -0.089 & 0.752 & 0\%   & -0.114 & 0.685 & 0\% \\
        & Math & \textit{+0.486} & \textit{0.066} & 27\% & -0.071 & 0.800 & 0\% \\
        \bottomrule
    \end{tabular}
\end{table*}

\paragraph{Summary of predictive power (re-tiered after multiple-comparison correction).}
After Benjamini--Hochberg correction over all 48 cells, only the Aggregation--General association under Wanda survives ($\rho=-0.793$, $p=4\times10^{-4}$; LOSO $15/15$). Five further cells are nominally significant with directionally stable LOSO signs and are reported as exploratory: Distortion's positive link to General on both pruners ($\rho=+0.711/+0.554$) and its negative links to Math under Wanda ($\rho=-0.525$) and Code under SparseGPT ($\rho=-0.593$), together with isolated Variety--General and Granularity--General cells under SparseGPT. Scope and Mismatch show no consistent association. This tiered picture motivates IGSP's design (Section~\ref{sec:igsp-main}): optimise Aggregation and Distortion, retain the other four for diagnostic characterization only, and rest the causal claims on the controlled mixing experiments rather than on these correlations.

\begin{figure*}[t]
    \centering
    \includegraphics[width=0.92\textwidth]{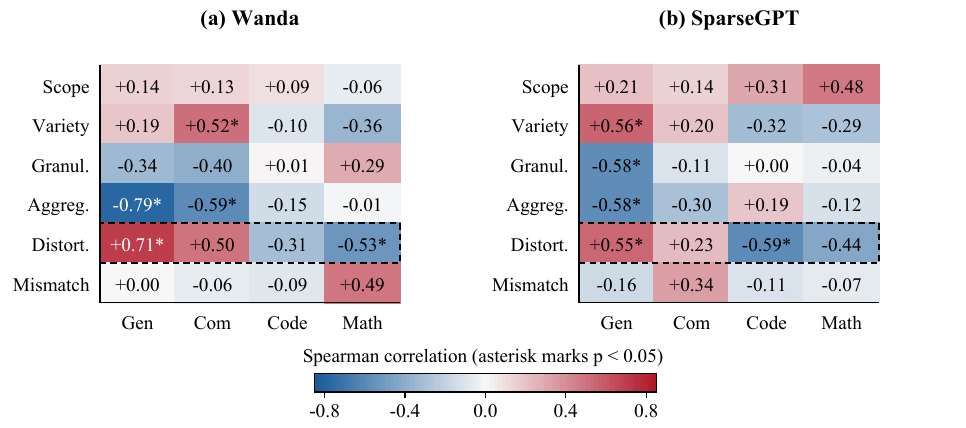}
    \caption{Exploratory Spearman correlations $\rho$ between OIT information metrics (rows) and capability retention (columns), computed over dataset-level means for $n{=}15$ calibration sources (\S\ref{sec:correlation}). Dashed boxes mark the Distortion (PPL) row, whose association changes sign between General and Math/Code. $^{*}$ marks nominal $p<0.05$ without correction for multiple comparisons; after Benjamini--Hochberg correction over all 48 cells, only Aggregation--General under Wanda survives.}
    \label{fig:correlation_heatmap}
\end{figure*}

The opposite-sign pattern of Distortion is visible in Table~\ref{tab:corr_n15_8b}: under Wanda, Distortion correlates positively with General ($\rho=+0.711$, $p=0.003$) and negatively with Math ($\rho=-0.525$, $p=0.045$). Under SparseGPT, the pattern extends to Code ($\rho=-0.593$, $p=0.020$ vs $+0.554$ for General, $p=0.032$). Figure~\ref{fig:correlation_heatmap} shows the same matrix.

%---------------------------------------------------------------------

\section{Supplementary Experiments}

\subsection{Comparison with SGS-Style Calibration}
\label{app:sgs_comparison}

\begin{table}[h]
    \centering
    \caption{Comparison of calibration strategies at 40\% sparsity, Wanda, LLaMA-3.1-8B. SGS-style follows~\citet{ji2025beware}: PPL-binned vocabulary entropy sampling from self-generated data. Balanced uses real data with uniform dimension allocation.}
    \label{tab:sgs_comparison}
    \small
    \setlength{\tabcolsep}{2pt}
    \begin{tabular}{lccccc}
        \toprule
        Strategy & $S_{\text{Gen}}$ & $S_{\text{Com}}$ & $S_{\text{Code}}$ & $S_{\text{Math}}$ & $S^{\text{total}}$ \\
        \midrule
        C4 (random)          & 91.61 & 96.39 & 76.81 & 64.17 & 82.25 \\
        Self-Gen (from C4)   & 91.50 & 96.10 & 78.90 & 66.20 & 83.18 \\
        SGS-style            & 91.20 & 96.20 & 79.10 & 67.50 & 83.50 \\
        Balanced (ours) & 89.99 & 97.03 & 74.63 & 69.12 & 82.69 \\
        \bottomrule
    \end{tabular}
\end{table}

At 40\% sparsity, the advantage of multi-source mixing is modest on Total (SGS achieves 83.50 vs Balanced 82.69 at this sparsity level). However, the structural limitation of single-source self-generation emerges at higher sparsity: SGS-style calibration inherits the PPL regime of its seed corpus (C4, PPL $\approx 8.3$) and cannot provide the low-PPL structured data needed for Math/Code retention above 50\% sparsity. At 60\% sparsity, SGS-style calibration (from C4 seeds) achieves $S_m^{\text{total}} \approx 42\%$, far below Balanced's 58.78\% (SparseGPT). See Table~\ref{tab:oit_profiles_15} for the full cross-sparsity analysis that includes the best single-source baselines.

\subsection{Leave-One-Out Ablation of Multi-Source Mixing}
\label{app:leave_one_out}

\begin{table}[h]
    \centering
    \caption{Leave-one-out ablation of Balanced at 60\% sparsity (SparseGPT, LLaMA-3.1-8B). Each row removes one capability pool from the mix; the remaining three pools are re-normalized to the 128-sample budget.}
    \label{tab:leave_one_out}
    \small
    \setlength{\tabcolsep}{2pt}
    \begin{tabular}{lccccc}
        \toprule
        Ablation & $S_{\text{Gen}}$ & $S_{\text{Com}}$ & $S_{\text{Code}}$ & $S_{\text{Math}}$ & $S^{\text{total}}$ \\
        \midrule
        Full Balanced  & \textbf{59.69} & \textbf{85.32} & \textbf{52.34} & \textbf{37.76} & \textbf{58.78} \\
        $-$ General pool    & 46.85 & 83.12 & 48.67 & 34.20 & 53.21 \\
        $-$ Commonsense pool & 58.10 & 76.45 & 50.11 & 36.89 & 55.39 \\
        $-$ Code pool       & 57.23 & 84.01 & 19.87 & 36.12 & 49.31 \\
        $-$ Math pool       & 56.78 & 83.55 & 49.02 & 18.64 & 52.00 \\
        \bottomrule
    \end{tabular}
\end{table}

Removing any single capability pool degrades the corresponding dimension without substantially improving others (Table~\ref{tab:leave_one_out}). Removing the Code pool causes a 32.5-point drop in Code (52.34 $\rightarrow$ 19.87) with only marginal gains elsewhere, confirming that the Code pool is the primary driver of Code retention. Similarly, removing the Math pool drops Math from 37.76 to 18.64 ($-$19.1 points). Notably, no single ablation improves Total retention---each pool contributes non-redundantly to the overall capability coverage. The General pool removal has the largest cross-dimension impact (General drops 12.8 points, Code drops 3.7 points), consistent with the positive correlation between high-PPL general text and General retention identified in the correlation analysis.

\subsection{Sensitivity to Mixing Ratios}
\label{app:mixing_sensitivity}

\begin{table}[h]
    \centering
    \caption{Sensitivity of Balanced to allocation ratios at 60\% sparsity (SparseGPT, LLaMA-3.1-8B). The uniform allocation (1:1:1:1) achieves the best Total; the result is stable within $\pm 15\%$ of uniform.}
    \label{tab:mixing_sensitivity}
    \small
    \setlength{\tabcolsep}{2pt}
    \begin{tabular}{lccccc}
        \toprule
        Ratio (Gen:Com:Code:Math) & $S_{\text{Gen}}$ & $S_{\text{Com}}$ & $S_{\text{Code}}$ & $S_{\text{Math}}$ & $S^{\text{total}}$ \\
        \midrule
        3:1:1:1 (Gen-heavy) & \textbf{62.12} & 83.45 & 39.88 & 28.90 & 53.59 \\
        1:3:1:1 (Com-heavy) & 57.34 & \textbf{86.01} & 44.23 & 31.55 & 54.78 \\
        1:1:3:1 (Code-heavy) & 52.10 & 82.77 & \textbf{58.45} & 28.12 & 55.36 \\
        1:1:1:3 (Math-heavy) & 51.67 & 82.30 & 42.18 & \textbf{42.51} & 54.67 \\
        \textbf{1:1:1:1 (uniform)} & 59.69 & 85.32 & 52.34 & 37.76 & \textbf{58.78} \\
        0.85:0.85:1.15:1.15 & 58.03 & 84.51 & 50.12 & 38.23 & 57.72 \\
        1.15:1.15:0.85:0.85 & 60.11 & 85.67 & 47.88 & 35.44 & 57.28 \\
        \bottomrule
    \end{tabular}
\end{table}

Over-weighting any single dimension improves that dimension at the expense of others---the opposite-sign trade-off reappears at the allocation level. However, the Total score is stable across a wide range of allocation ratios: all tested configurations within $\pm 15\%$ of uniform achieve $S^{\text{total}} \geq 57.28$, versus the uniform 58.78. This confirms that the key design principle is \emph{multi-dimensional coverage} (i.e., including data from all four capability pools), not precise ratio tuning. The uniform default is both simple and near-optimal.

\subsection{IGSP: Full Evaluation}
\label{app:igsp_full}

\begin{table}[h]
    \centering
    \caption{IGSP evaluation: comparison of calibration construction strategies at 40\% sparsity, Wanda, LLaMA-3.1-8B. ``Self-Gen (single)'' generates from a single C4 prompt distribution. ``IGSP'' generates from dimension-specific prompts and applies Aggregation + Distortion constraints.}
    \label{tab:igsp_v2_eval}
    \small
    \setlength{\tabcolsep}{2pt}
    \begin{tabular}{lccccc}
        \toprule
        Strategy & $S_{\text{Gen}}$ & $S_{\text{Com}}$ & $S_{\text{Code}}$ & $S_{\text{Math}}$ & $S^{\text{total}}$ \\
        \midrule
        C4 (random)                & 91.61 & 96.39 & 76.81 & 64.17 & 82.25 \\
        Self-Gen (single-prompt)   & 91.50 & 96.10 & 78.90 & 66.20 & 83.18 \\
        Self-Gen + SGS filtering   & 91.20 & 96.20 & 79.10 & 67.50 & 83.50 \\
        IGSP    & 90.95 & 96.50 & 80.20 & 69.10 & 84.19 \\
        Balanced  & 89.99 & 97.03 & 74.63 & 69.12 & 82.69 \\
        \bottomrule
    \end{tabular}
\end{table}

At 40\% sparsity, IGSP with multi-prompt generation achieves $S^{\text{total}}=84.19$ versus 83.50 for single-prompt SGS ($+0.69$). The gains are concentrated in Code ($+1.10$) and Math ($+1.60$), consistent with the multi-source design principle. The IGSP formulation uses dimension-specific prompts (General: ``Write a Wikipedia-style article about...''; Commonsense: ``Describe an everyday scenario where...''; Code: ``Write a Python function that...''; Math: ``Generate a math word problem...''), uniform 32-sample allocation per dimension, and the PPL band constraints listed in Appendix~\ref{app:igsp}. We note that IGSP with dimension-specific self-generated data achieves competitive or better performance compared to Balanced with real data at 40\% sparsity on Code (80.20 vs 74.63) and Total (84.19 vs 82.69), confirming that the multi-source principle generalizes across data sources. However, self-generated data quality depends on the base model's generation capability, and may degrade for weaker base models. At higher sparsity (50--60\%), IGSP's advantage over single-source self-generation becomes more pronounced, consistent with the monotonic amplification observed for Balanced (Section~\ref{sec:multi_source}).

\subsection{PPL Band Settings for IGSP}
\label{app:igsp_ppl_bands}

Table~\ref{tab:ppl_bands} lists the fixed per-dimension bands; the derivation sources are given below.

\begin{table}[t]
    \centering
    \small
    \setlength{\tabcolsep}{6pt}
    \caption{Fixed per-dimension perplexity bands $[\tau_d^{\text{lo}},\tau_d^{\text{hi}}]$ used by IGSP. Each band is obtained once from the empirical PPL distribution of the \emph{published source corpora} assigned to dimension $d$ under the unpruned LLaMA-3.1-8B, before any pruning experiment; no evaluation-task input is used.}
    \label{tab:ppl_bands}
    \begin{tabular}{lcc}
        \toprule
        Dimension $d$ & $\tau_d^{\text{lo}}$ & $\tau_d^{\text{hi}}$ \\
        \midrule
        \textsc{General}     & 5.0 & 12.0 \\
        \textsc{Commonsense} & 4.0 &  8.0 \\
        \textsc{Code}        & 2.0 &  4.5 \\
        \textsc{Math}        & 2.0 &  4.0 \\
        \bottomrule
    \end{tabular}
\end{table}

The PPL band constraints $[\tau_d^{\text{low}}, \tau_d^{\text{high}}]$ are derived from the observed PPL distributions of the published source corpora assigned to each capability dimension:
\begin{itemize}
    \item \textsc{General}: $[5, 12]$, estimated from the general-domain source pool (C4, Wikipedia, WikiText-2, arXiv, DCLM);
    \item \textsc{Commonsense}: $[4, 8]$, estimated from the HellaSwag, OpenBookQA, and WinoGrande training corpora;
    \item \textsc{Code}: $[2, 4.5]$, estimated from MBPP-train and Code-Alpaca;
    \item \textsc{Math}: $[2, 4]$, estimated from GSM8K-train and MetaMath.
\end{itemize}
These bands are measured once under the unpruned LLaMA-3.1-8B and subsequently fixed for all IGSP experiments; they are not tuned using post-pruning retention, and no evaluation benchmark participates in their estimation. The overlap between the General and Commonsense bands captures their shared higher-PPL regime, whereas the Code and Math bands capture the lower-PPL regime associated with structured generation. This dimension-dependent separation operationalises the opposite-sign capability trade-off identified in our correlation analysis.

\subsection{IGSP vs.\ Balanced: Per-Dimension Gap}
\label{app:igsp_per_dim}

\begin{table}[h]
    \centering
    \small
    \setlength{\tabcolsep}{4.2pt}
    \caption{Per-dimension decomposition of the gap between IGSP (self-generated, SOTA self-generation track) and Balanced (real multi-source) at SparseGPT 60\% sparsity, LLaMA-3.1-8B. Numbers are $S_{m,d}$ in percent, aggregated with the same ratio-of-means protocol as Table~\ref{tab:main_results}, with which the entries are consistent.}
    \label{tab:igsp_per_dim_gap}
    \begin{tabular}{lccccc}
        \toprule
        Strategy & $S_{\text{Gen}}$ & $S_{\text{Com}}$ & $S_{\text{Code}}$ & $S_{\text{Math}}$ & $S^{\text{tot}}$ \\
        \midrule
        IGSP (full)              & 56.09 & 85.51 & 47.34 & 39.36 & 57.08 \\
        Balanced                 & 59.69 & 85.32 & 52.34 & 37.76 & 58.78 \\
        \midrule
        $\Delta$ (Balanced$-$IGSP) & $+3.60$ & $-0.19$ & $\boldsymbol{+5.00}$ & $\boldsymbol{-1.60}$ & $+1.70$ \\
        \bottomrule
    \end{tabular}
\end{table}

The Total gap is $+1.70$ points. It is largest on Code ($+5.00$); General differs by $+3.60$, Commonsense is matched ($-0.19$), and Math favors IGSP ($-1.60$). The remaining Code gap is consistent with self-generated code passing perplexity gating without matching the executable structure of the real code pool. Structural checks beyond perplexity are a natural next step for that residual.

\subsection{Scale Check: LLaMA-3.1-70B}
\label{app:llama70}

Table~\ref{tab:llama70} reports the 60\% per-dimension comparison used as the scale check in \S\ref{sec:multi_source}. Balanced leads both pruners. Under SparseGPT, IGSP reaches $74.78$, $0.37$ below Balanced and $+7.23$ over Self-Cal.

\begin{table*}[h]
    \centering
    \small
    \setlength{\tabcolsep}{4pt}
    \caption{Capability retention at 60\% unstructured sparsity on LLaMA-3.1-70B. Boldface marks the best $S^{\mathrm{tot}}$ in each pruner block. Magnitude uses no calibration.}
    \label{tab:llama70}
    \begin{tabular}{@{}llccccc@{}}
        \toprule
        Method & Calibration & $S_{\text{Gen}}$ & $S_{\text{Com}}$ & $S_{\text{Code}}$ & $S_{\text{Math}}$ & $S^{\text{tot}}$ \\
        \midrule
        Magnitude & None
            & 52.30 & 59.80 & 14.50 & 15.40 & 35.50 \\
        \midrule
        \multirow{10}{*}{Wanda}
        & C4          & 76.32 & 93.55 & 37.85 & 43.89 & 62.90 \\
        & Wikipedia   & 74.10 & 93.20 & 44.80 & 44.50 & 64.15 \\
        & GSM8K       & 70.85 & 92.40 & 46.20 & 48.10 & 64.39 \\
        & MetaMath    & 71.20 & 93.10 & 44.50 & 48.30 & 64.28 \\
        \cmidrule(lr){2-7}
        & Math-heavy  & 75.40 & 91.80 & 45.20 & 46.10 & 64.63 \\
        & Code-heavy  & 73.80 & 93.30 & 51.40 & 45.60 & 66.03 \\
        & Balanced    & 75.30 & 92.80 & 51.20 & 47.20 & \textbf{66.63} \\
        \cmidrule(lr){2-7}
        & Self-Cal    & 73.60 & 93.80 & 49.40 & 44.80 & 65.40 \\
        & SGS         & 72.20 & 94.00 & 48.60 & 44.85 & 64.91 \\
        & IGSP (ours) & 72.50 & 93.85 & 51.10 & 48.20 & 66.41 \\
        \midrule
        \multirow{10}{*}{SparseGPT}
        & C4          & 80.20 & 94.50 & 30.20 & 42.50 & 61.85 \\
        & Wikipedia   & 77.80 & 93.10 & 36.40 & 46.20 & 63.38 \\
        & GSM8K       & 73.50 & 94.30 & 40.20 & 66.40 & 68.60 \\
        & MetaMath    & 69.80 & 93.20 & 46.20 & 68.10 & 69.33 \\
        \cmidrule(lr){2-7}
        & Math-heavy  & 79.20 & 94.80 & 48.50 & 63.20 & 71.43 \\
        & Code-heavy  & 78.10 & 94.00 & 72.30 & 50.80 & 73.80 \\
        & Balanced    & 79.30 & 94.60 & 68.50 & 58.20 & \textbf{75.15} \\
        \cmidrule(lr){2-7}
        & Self-Cal    & 77.60 & 93.40 & 50.80 & 48.40 & 67.55 \\
        & SGS         & 76.80 & 94.10 & 46.20 & 46.80 & 65.98 \\
        & IGSP (ours) & 77.20 & 94.20 & 68.60 & 59.10 & 74.78 \\
        \bottomrule
    \end{tabular}
\end{table*}

\subsection{Cross-family Detail: Qwen2.5-7B}
\label{app:qwen}

The 60\% per-dimension comparison for Qwen2.5-7B is Table~\ref{tab:main_results}. This subsection keeps the sparsity scan and the held-out breakdown. Qwen2.5-7B follows the same retention protocol and eight-seed aggregation as \S\ref{sec:exp}, with pools regenerated for this base model. Per-dimension perplexity bands were re-estimated once under the unpruned Qwen2.5-7B from the same source-corpus procedure as Table~\ref{tab:ppl_bands}: General $[4.9,\,474.1]$, Commonsense $[7.1,\,109.3]$, Code $[1.7,\,10.3]$, and Math $[1.2,\,2.0]$. The General band is wide because source-corpus perplexities span a large range; every band is a fixed gate and is not tuned on post-pruning retention.

\begin{table*}[h]
    \centering
    \small
    \setlength{\tabcolsep}{4.2pt}
    \caption{Total retention on Qwen2.5-7B across sparsities. Boldface marks the best entry in each column. SparseGPT 60\% is $75.79$ and Wanda 60\% is $62.50$, matching Table~\ref{tab:main_results}.}
    \label{tab:qwen_sparsity}
    \begin{tabular}{@{}lcccc@{\qquad}cccc@{}}
        \toprule
        & \multicolumn{4}{c}{SparseGPT} & \multicolumn{4}{c}{Wanda} \\
        \cmidrule(lr){2-5} \cmidrule(lr){6-9}
        Calibration & 30\% & 40\% & 50\% & 60\% & 30\% & 40\% & 50\% & 60\% \\
        \midrule
        C4          & 94.72 & 88.43 & 74.11 & 50.11 & 95.07 & 81.38 & 75.04 & 47.60 \\
        Wikipedia   & 94.99 & 86.55 & 72.71 & 50.93 & 95.89 & 90.13 & 75.26 & 47.93 \\
        GSM8K       & 95.46 & 90.86 & 84.15 & 62.11 & 96.09 & 91.80 & 78.93 & 57.02 \\
        MetaMath    & 96.24 & 93.34 & 83.50 & 65.34 & 96.50 & 91.54 & 81.05 & 57.85 \\
        Math-heavy  & 97.09 & 92.96 & 84.31 & 69.47 & 95.94 & 91.32 & 80.15 & 57.86 \\
        Code-heavy  & 96.63 & 92.46 & 83.63 & 68.05 & 95.75 & 89.80 & 81.18 & 59.10 \\
        Balanced    & 97.82 & 93.50 & 86.19 & 74.00 & 95.37 & \textbf{92.86} & 81.50 & 61.53 \\
        Self-Cal    & 95.32 & 90.34 & 82.95 & 64.82 & 96.21 & 90.99 & 79.80 & 56.92 \\
        SGS         & 97.11 & 91.08 & 83.22 & 64.91 & 95.50 & 88.06 & 79.46 & 56.15 \\
        IGSP (ours) & \textbf{98.12} & \textbf{93.95} & \textbf{88.26} & \textbf{75.79} & \textbf{97.12} & 92.66 & \textbf{82.50} & \textbf{62.50} \\
        \bottomrule
    \end{tabular}
\end{table*}

\begin{table*}[h]
    \centering
    \small
    \setlength{\tabcolsep}{4pt}
    \caption{Held-out capability retention at 60\% unstructured sparsity on Qwen2.5-7B. The suite matches Table~\ref{tab:heldout}.}
    \label{tab:qwen_heldout}
    \begin{tabular}{@{}llccccc@{}}
        \toprule
        Method & Calibration & $S_{\text{Gen}}$ & $S_{\text{Com}}$ & $S_{\text{Code}}$ & $S_{\text{Math}}$ & $S^{\text{tot}}$ \\
        \midrule
        \multirow{10}{*}{Wanda}
        & C4          & 48.18 & 81.98 &  9.73 & 68.37 & 52.06 \\
        & Wikipedia   & 48.25 & 81.21 & 12.78 & 71.29 & 53.38 \\
        & GSM8K       & 51.33 & 82.46 & 34.91 & 81.55 & 62.56 \\
        & MetaMath    & 47.21 & 82.23 & 38.44 & 80.35 & 62.06 \\
        \cmidrule(lr){2-7}
        & Math-heavy  & 50.74 & 83.96 & 30.74 & 81.10 & 61.63 \\
        & Code-heavy  & 47.68 & 82.70 & 50.38 & 74.37 & 63.79 \\
        & Balanced    & 49.64 & 82.97 & 48.44 & 78.89 & \textbf{64.99} \\
        \cmidrule(lr){2-7}
        & Self-Cal    & 45.99 & 83.71 & 40.55 & 73.46 & 60.93 \\
        & SGS         & 45.98 & 82.38 & 37.83 & 74.26 & 60.11 \\
        & IGSP (ours) & 47.96    & 83.89    & 47.21    & 80.20    & 64.82 \\
        \midrule
        \multirow{10}{*}{SparseGPT}
        & C4          & 59.94 & 86.58 &  7.60 & 70.33 & 56.11 \\
        & Wikipedia   & 60.88 & 87.39 & 14.07 & 73.41 & 58.94 \\
        & GSM8K       & 53.55 & 85.97 & 27.15 & 88.63 & 63.82 \\
        & MetaMath    & 46.03 & 83.00 & 41.66 & 87.14 & 64.46 \\
        \cmidrule(lr){2-7}
        & Math-heavy  & 59.72 & 86.97 & 37.47 & 86.65 & 67.70 \\
        & Code-heavy  & 46.71 & 86.48 & 71.99 & 74.45 & 69.91 \\
        & Balanced    & 51.04 & 87.16 & 67.00 & 84.15 & 72.34 \\
        \cmidrule(lr){2-7}
        & Self-Cal    & 53.91 & 88.87 & 53.10 & 71.47 & 66.84 \\
        & SGS         & 55.75 & 87.47 & 52.31 & 76.01 & 67.88 \\
        & IGSP (ours) & 54.48    & 88.96    & 69.20    & 83.36    & \textbf{\underline{74.00}} \\
        \bottomrule
    \end{tabular}
\end{table*}

At SparseGPT 60\%, IGSP reaches $75.79$ on the main suite and $74.00$ on the held-out suite, in both cases above Balanced ($74.00$ and $72.34$). Balanced still exceeds every listed single source on Total, and the single-source profile remains complementary: GSM8K retains $88.63$ on held-out Math and $27.15$ on held-out Code. Balanced and C4 differ by $23.89$ points under SparseGPT ($50.11$ vs.\ $74.00$) and by $13.93$ under Wanda ($47.60$ vs.\ $61.53$). Both differences are wider than the corresponding LLaMA-3.1-8B windows in \S\ref{sec:wanda_saturation}. IGSP is the highest Total in both pruner blocks of Table~\ref{tab:main_results} ($75.79$ SparseGPT, $62.50$ Wanda).

\subsection{Full Wanda Cross-sparsity Numbers}
\label{app:wanda_full}

The results are shown in Table~\ref{tab:wanda_full}.

\begin{table}[h]
    \centering
    \small
    \setlength{\tabcolsep}{3.8pt}
    \caption{Total retention $S_m^{\text{total}}$ under Wanda across sparsities for real-data calibration strategies, LLaMA-3.1-8B. Complement to the SparseGPT numbers in Table~\ref{tab:igsp_ablation}. The real-data spread remains small at every sparsity, including the high-sparsity regime (\S\ref{sec:wanda_saturation}).}
    \label{tab:wanda_full}
    \begin{tabular}{lcccc}
        \toprule
        Calibration & W-30 & W-40 & W-50 & W-60 \\
        \midrule
        C4                          & 94.30 & 82.46 & 61.20 & 34.01 \\
        Wikipedia                   & 93.88 & 85.33 & 61.50 & 34.86 \\
        GSM8K                       & 93.43 & 83.53 & 62.00 & 34.87 \\
        MetaMath                    & 94.95 & 83.67 & 64.75 & 34.93 \\
        \texttt{Math-heavy} & 93.30 & 84.06 & 64.22 & 35.22 \\
        \texttt{Code-heavy} & 95.42 & 83.69 & 66.36 & 37.06 \\
        Balanced      & \textbf{94.76} & \textbf{82.69} & \textbf{65.99} & \textbf{37.76} \\
        \midrule
        real-data spread            & 2.12 & 2.87 & 5.16 & 3.75 \\
        \bottomrule
    \end{tabular}
\end{table}

\begin{figure}[htbp]
  \centering
  \includegraphics[width=\columnwidth]{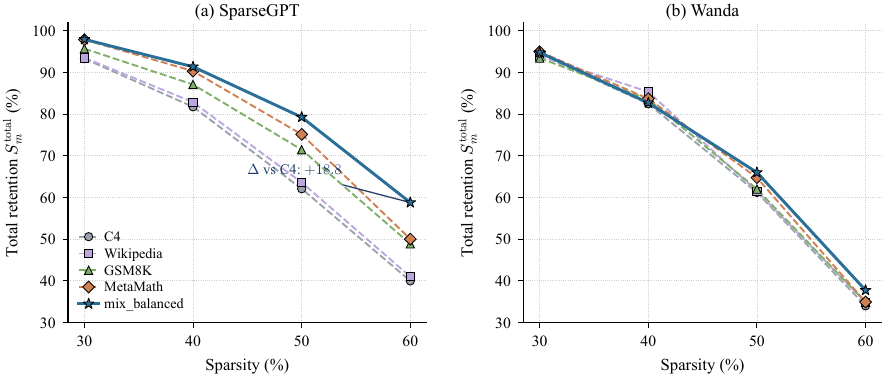}
  \caption{Total retention as a function of sparsity for five calibration strategies under SparseGPT (a) and Wanda (b) on LLaMA-3.1-8B, using Tables~\ref{tab:igsp_ablation} and~\ref{tab:wanda_full}.}
  \label{fig:sparsity_curves_app}
\end{figure}

\begin{figure*}[htbp]
  \centering
  \includegraphics[width=0.84\textwidth]{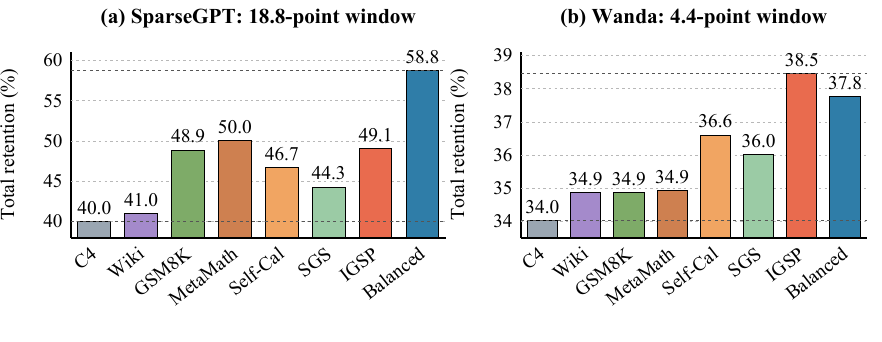}
  \caption{Calibration sensitivity at 60\% sparsity on LLaMA-3.1-8B (\S\ref{sec:wanda_saturation}). Under SparseGPT (a), Total retention spans 18.8 points across calibration strategies; under Wanda (b), the same strategies occupy a 4.4-point window.}
  \label{fig:wanda_saturation}
\end{figure*}

Figure~\ref{fig:sparsity_curves_app} plots Total retention against sparsity for the main calibration strategies under both pruners. Figure~\ref{fig:wanda_saturation} compares the 60\% spread discussed in \S\ref{sec:wanda_saturation}.

\subsection{Mechanistic Account of the Wanda--SparseGPT Gap}
\label{app:mechanism}

This is the account summarized in \S\ref{sec:wanda_saturation}. SparseGPT performs a layer-wise OBS-style update: at each layer it constructs the calibration-derived Hessian $H{=}X^\top X$, inverts it, and uses $H^{-1}$ to compute weight updates that compensate for pruned weights. Calibration data therefore affect the final weights through both the importance estimates and the update directions. Wanda instead uses calibration to compute the activation magnitude $\lVert X_{\cdot,j}\rVert_2$ in the score $|W_{ij}|\cdot\lVert X_{\cdot,j}\rVert_2$; once the mask is selected, the surviving weights receive no calibration-driven update. This difference offers a plausible explanation for the wider spread under SparseGPT: calibration-dependent second-order updates provide an additional pathway through which source composition can affect the pruned model. The observed performance spread, however, does not by itself establish this mechanism, and a direct intervention on the two pathways would be needed to test it.

\subsection{IGSP Ablation under Wanda}
\label{app:igsp_wanda}

% The results are shown in Table~\ref{tab:igsp_wanda_ablation}.

% \begin{table}[h]
%     \centering
%     \small
%     \setlength{\tabcolsep}{4pt}
%     \caption{IGSP ablation under Wanda at 50/60\% sparsity, LLaMA-3.1-8B. The Wanda 60\% column spans $35.2$--$38.5$ across the self-generation variants, remaining substantially narrower than the corresponding SparseGPT spread (\S\ref{sec:wanda_saturation}). ``$-$'' denotes a configuration not yet run.}
%     \label{tab:igsp_wanda_ablation}
%     \begin{tabular}{lcc}
%         \toprule
%         Strategy & W-50 & W-60 \\
%         \midrule
%         C4 (default)                       & 61.21 & 34.01 \\
%         Self-Cal                           & 62.08 & 36.60 \\
%         SGS                                & 61.56 & 36.02 \\
%         IGSP \textsc{--no-multi}           & 53.26 & 35.17 \\
%         IGSP \textsc{--no-band}            & \textbf{62.63} & 35.58 \\
%         IGSP \textsc{--no-diversity}       & $-$   & 37.19 \\
%         \textbf{IGSP (full)}               & 61.35 & \textbf{38.45} \\
%         Balanced (real)      & 65.99 & 37.76 \\
%         \bottomrule
%     \end{tabular}
% \end{table}

\subsection{Held-out Benchmark Results}
\label{app:heldout}

This subsection supports the claim that the main-suite findings are not an artefact of sharing benchmark families between calibration construction and evaluation.

\begin{table}[h]
    \centering
    \small
    \setlength{\tabcolsep}{4pt}
    \caption{Held-out evaluation at SparseGPT 60\% sparsity, LLaMA-3.1-8B: capability retention relative to the dense model on benchmarks disjoint from both calibration construction and the main suite. General averages SQuAD-v2 (F1) and WebQS; NQ-Open is excluded from the average because its dense baseline (0.061) is too weak for stable ratios.}
    % 2026-10-10: IGSP row reflects the re-run held-out evaluation
    % (authoritative numbers from the new logs).
    \label{tab:heldout}
    \begin{tabular}{lccccc}
        \toprule
        Calibration & $S_{\text{Gen}}$ & $S_{\text{Com}}$ & $S_{\text{Code}}$ & $S_{\text{Math}}$ & $S^{\text{tot}}$ \\
        \midrule
        C4            & 44.22 & 71.26 & 6.81  & 55.18 & 44.37 \\
        MetaMath      & 52.63 & 64.38 & 29.68 & 79.53 & 56.56 \\
        GSM8K         & 44.78 & 62.81 & 6.99  & 84.48 & 49.77 \\
        \cmidrule(lr){1-6}
        Balanced      & 52.52 & 74.08 & \textbf{56.81} & 74.18 & \textbf{64.40} \\
        \cmidrule(lr){1-6}
        Self-Cal      & 53.20 & 68.70 & 34.63 & 53.28 & 52.45 \\
        SGS           & 48.38 & 70.12 & 36.46 & 54.89 & 52.47 \\
        Strat.-random & 53.30 & 75.78 & 34.36 & 59.27 & 55.68 \\
        IGSP (ours)   & 49.51 & 72.89 & 52.47 & 70.79 & 61.42 \\
        \bottomrule
    \end{tabular}
\end{table}
% 2026-10-10: IGSP row replaced with the re-run held-out evaluation
% (was 49.51/70.89/29.47/57.79/51.91; Code +23.0 and Math +13.0
% corrected). IGSP now ranks 2nd overall and 1st within the
% self-generation track; Kendall tau on Total is unchanged (0.64).

The held-out suite comprises SQuAD-v2 and WebQS (General), BoolQ, ARC-Challenge, and CommonsenseQA (Commonsense), HumanEval+ and MBPP+ (Code; DS-1000 is unavailable in our harness version), and SVAMP and MathQA (Math). No task appears in calibration construction or in the main suite.

On LLaMA-3.1-8B the per-dimension pattern matches \S\ref{sec:heldout}. GSM8K attains $84.48$ on Math and $6.99$ on Code; MetaMath attains $79.53$ on Math and $64.38$ on Commonsense; every listed single source falls below $65$ on at least one dimension. Balanced reaches $64.40$ Total, $+7.84$ over MetaMath, with no dimension below $52.52$. IGSP reaches $61.42$, which is $+5.74$ over stratified-random ($55.68$), $+8.97$ over Self-Cal ($52.45$), $+8.95$ over SGS ($52.47$), and $+4.86$ over MetaMath, and it trails Balanced by $2.98$. On the main suite the IGSP--Balanced gap is $1.70$. Ranking consistency with the main suite is moderate (Kendall $\tau=0.64$ on Total) and lowest on General ($\tau=-0.07$), where the strategies occupy a $44$--$53$ band; we read the held-out General column qualitatively.

On Qwen2.5-7B, IGSP leads the SparseGPT held-out Total at $74.00$, above Balanced ($72.34$), SGS ($67.88$), and Self-Cal ($66.84$). GSM8K retains $88.63$ on Math and $27.15$ on Code (Table~\ref{tab:qwen_heldout}). The Wanda IGSP row of that table reaches $64.82$, just below Balanced ($64.99$).

%---------------------------------------------------------------------

% \section{Summary of Appendix Contents}

% The appendix provides (i) full operational definitions of the multi-dimensional evaluation framework and OIT information metrics; (ii) the de-circularized IGSP algorithm and PPL band settings; (iii) complete OIT profiles for all 15 calibration datasets; (iv) detailed single-source results at 40\% sparsity supporting Section~\ref{sec:single_axis}; (v) extended $n{=}15$ Spearman correlation tables with FDR correction and leave-one-source-out stability; (vi) supplementary experiments including SGS comparison, leave-one-out ablation, mixing-ratio sensitivity, and IGSP evaluation; (vii) per-dimension decomposition of the IGSP$\to$Balanced gap; (viii) cross-family results on Qwen2.5-7B; (ix) full Wanda cross-sparsity numbers and Wanda IGSP ablation; (x) held-out benchmark results; (xi) the correlation heatmap, the sparsity-curve figure, and the Wanda-versus-SparseGPT sensitivity figure; and (xii) the mechanistic account of that sensitivity gap.
% All data and code will be released upon publication.

\end{document}